%% file: main.tex
\documentclass[9.5pt,journal,compsoc]{IEEEtran}

\usepackage{amsmath,amssymb,amsfonts}
\usepackage{graphicx}
\usepackage{textcomp}
\usepackage{subcaption}
\usepackage{xcolor}
\def\BibTeX{{\rm B\kern-.05em{\sc i\kern-.025em b}\kern-.08em
		T\kern-.1667em\lower.7ex\hbox{E}\kern-.125emX}}

\input{includes}

\ifCLASSOPTIONcompsoc
\usepackage[nocompress]{cite}
\else
\usepackage{cite}
\fi

\ifCLASSINFOpdf
\else
\fi

\begin{document}
	\title{An Experimental Study of Byzantine-Robust Aggregation Schemes in Federated Learning}

	\author{Shenghui~Li,~\IEEEmembership{Student~Member,~IEEE,}
		Edith~C.-H.~Ngai,~\IEEEmembership{Senior~Member,~IEEE,}
		and~Thiemo~Voigt,~\IEEEmembership{Member,~IEEE}
		\thanks{This research was supported by the RGC General Research Funds No. 17203320 and No. 17209822 from Hong Kong, the Swedish Research Council project grant No. 2017-04543, HKU-TCL joint research centre for artificial intelligence seed funding, and the European Union’s Horizon 2020 research and innovation programme under grant agreement No. 101015922.}
		\thanks{Shenghui~Li is with the Department of Information Technology, Uppsala University, Uppsala, Sweden~(e-mail: shenghui.li@it.uu.se).}
		\thanks{Edith~C.-H.~Ngai (corresponding author) is with the Department of Electrical and Electronic Engineering, The University of Hong Kong, Hong Kong, China~(e-mail: chngai@eee.hku.hk).}
		\thanks{Thiemo~Voigt is with the Department of Electrical Engineering, Uppsala University, Uppsala, Sweden, and also with RISE, the Research Institutes of Sweden, Stockholm, Sweden~(e-mail: thiemo.voigt@it.uu.se).}
	}
	
	\markboth{Journal of \LaTeX\ Class Files,~Vol.~14, No.~8, August~2015}%
	{Shell \MakeLowercase{\textit{et al.}}: Bare Demo of IEEEtran.cls for Computer Society Journals}

	\IEEEtitleabstractindextext{%
		\begin{abstract}
			Byzantine-robust federated learning aims at mitigating Byzantine failures during the federated training process, where malicious participants (known as Byzantine clients) may upload arbitrary local updates to the central server in order to degrade the performance of the global model. In recent years, several robust aggregation schemes have been proposed to defend against malicious updates from Byzantine clients and improve the robustness of federated learning. These solutions were claimed to be Byzantine-robust, 
			under certain assumptions. Other than that, new attack strategies are emerging, striving to circumvent the defense schemes. However, there is a lack of systematical comparison and empirical study thereof. In this paper, we conduct an experimental study of Byzantine-robust aggregation schemes under different attacks using two popular algorithms in federated learning, \fedsgd and \fedavg. We first survey existing Byzantine attack strategies, as well as Byzantine-robust aggregation schemes that aim to defend against Byzantine attacks. We also propose a new scheme, \ours, to enhance the robustness of a clustering-based scheme by automatically clipping the updates. Then we provide an experimental evaluation of eight aggregation schemes in the scenario of five different Byzantine attacks. Our experimental results show that these aggregation schemes sustain relatively high accuracy in some cases, but they are not effective in all cases. In particular, our proposed \ours successfully defends against most attacks under independent and identically distributed (IID) local datasets. However, when the local datasets are Non-IID, the performance of all the aggregation schemes significantly decreases. With Non-IID data, some of these aggregation schemes fail even in the complete absence of Byzantine clients. Based on our experimental study, we conclude that the robustness of all the aggregation schemes is limited, highlighting the need for new defense strategies, in particular for Non-IID datasets.
		\end{abstract}
		
		\begin{IEEEkeywords}
			Byzantine attacks, distributed learning, federated learning,  neural networks, robustness
	\end{IEEEkeywords}}
	
	\maketitle

	\IEEEdisplaynontitleabstractindextext

	%
	\IEEEpeerreviewmaketitle

	\IEEEraisesectionheading{\section{Introduction}\label{sec:introduction}}

	%
	%
	%
	%
	\IEEEPARstart{F}{ederated} learning (FL)~\cite{mcmahan2017communication, yang2019federated,konevcny2016federated} is a machine learning paradigm for distributed model training on decentralized data across a set of client devices (e.g., desktops, mobile phones, IoT devices). Specifically, FL repeatedly performs the following steps: the server broadcasts the current global model to client devices; the clients then perform one or several local steps of stochastic gradient descent (SGD) using private training sets, and send the updates back to the server; then the server generates a new global model by aggregating the local updates to enable the next round of training. This paradigm allows the client devices to perform most of the computation, without requiring any of the participants to reveal their private training data to a centralized entity or each other. In addition, benefiting from multiple steps of local updates before uploading local updates, FL improves communication efficiency compared to traditional distributed learning~\cite{arjevani2015communication}.
	
	Despite the achievements of FL in terms of data privacy and communication efficiency, it also opens up the parameter updating process to manipulation by the clients, which brings serious security threats to  model training~\cite{lyu2020threats}. An important class of security threats in this context is known as Byzantine failures~\cite{lyu2020threats}, where some of the participants are not rigorously following the protocol, but upload arbitrary parameters to the central server, for example, due to faulty communication~\cite{ang2020robust}, or even worse, adversaries, where malicious attackers modify the update vectors to their desire and upload them to the server~\cite{fang2020local}. We use the term "Byzantine attack" to refer to the attacks where malicious attackers upload arbitrary updates to the server in order to degrade the overall performance of the global model in FL. In typical FL algorithms (e.g., \fedavg)~\cite{mcmahan2017communication}, the server aggregates the uploaded updates by calculating their sample mean and adds the result to the global model. However, it is well-known that the result of such an aggregation scheme can be arbitrarily skewed even by a single Byzantine client~\cite{yin2018byzantine}. The server thus requires Byzantine-robust solutions to defend against malicious clients. 
	
	In recent years, a number of Byzantine-robust techniques have been proposed~\cite{hu2021challenges}. They can be classified into three categories: redundancy-based schemes that assign each client redundant updates and use this redundancy to eliminate the effect of Byzantine failures~\cite{chen2018draco,rajput2019detox,sohn2020election,li2019rsa}; trust-based schemes that assume some of the clients or datasets are trusted for filtering and re-weighting the local model updates\cite{konstantinov2019robust, park2021sageflow,cao2021fltrust}; robust aggregation schemes that estimate the updates according to some robust aggregation algorithms~\cite{blanchard2017machine,yin2018byzantine,chen2017distributed,li2021byzantine,karimireddy2021learning,sattler2020byzantine}. 
	For the first category, redundancy-based schemes, in the worst case, require each node to compute $\Omega(M)$ times more updates, where $M$ is the number of Byzantine clients~\cite{chen2018draco}. This overhead is prohibitive in settings with large numbers of Byzantine clients. For the second category, the trusted clients/datasets are not always available to the server due to the concern of user data privacy. 
	
	Robust aggregation schemes, in contrast, aggregate the updates efficiently, without requiring trusted clients or datasets. However, typical schemes, including \gm~\cite{chen2017distributed}, \krum~\cite{blanchard2017machine}, \tm~\cite{yin2018byzantine}, \median~\cite{yin2018byzantine} and \cc~\cite{karimireddy2021learning}, often come with limited guarantees of Byzantine robustness (e.g., only establishing convergence to a limit, or only guaranteeing that the output of the aggregation scheme has a positive inner product with the true gradient~\cite{blanchard2017machine,guerraoui2018hidden}) and often require other strong assumptions, such as bounded absolute skewness~\cite{yin2018byzantine}. More importantly, recent studies reveal the vulnerability of some schemes to new attacks. For instance, the A Little Is Enough (ALIE) attack can circumvent \tm and \krum by taking advantage of empirical variance between the updates of clients if such variance is high enough~\cite{baruch2019little}. The Inner Product Manipulation (IPM) attack poses a significant threat to \median and \krum by manipulating the inner product between the true gradient and the robust aggregated gradients to be negative~\cite{xie2020fall}. Other schemes, such as \autogm~\cite{li2021byzantine} and \clustering~\cite{sattler2020byzantine}, were proposed with only empirical evaluations.
	
	These existing aggregation schemes are evaluated using different datasets, attack types, and hyper-parameters. There is a lack of empirical studies that compare different schemes of utilizing the same settings. Furthermore, the impact of data heterogeneity on robustness schemes is rarely evaluated as those schemes usually assume that all clients' local data are independent and identically distributed (IID). Therefore, there is a clear need for a comparative experimental study that offers in-depth insight into the performance of the existing Byzantine-robust schemes for FL.
	
	To meet this need, we conduct an experimental study on the Byzantine attack and defense problem in FL based on two well-known algorithms, \fedsgd and \fedavg~\cite{baruch2019little,mcmahan2017communication}. We first survey existing attack strategies and robust aggregation schemes in the literature. We further propose a new aggregation scheme \ours to address the weakness of an existing clustering-based scheme. Then we design experiments to evaluate the robustness of eight representative Byzantine-robust aggregation rules by applying five state-of-the-art attacking strategies. Our experimental results show that those aggregation rules sustain relatively high accuracy in some cases.  However, they are not effective in all cases. Moreover, when the local datasets are not independent and identically distributed (Non-IID), the capability of all the aggregation rules decreases significantly. With Non-IID data, some of these aggregation rules fail even in the complete absence of Byzantine clients. Furthermore, our proposed scheme performs the best in most attack scenarios when the datasets are IID. From the evaluation, we conclude that existing aggregation rules are insufficient to meet the need for Byzantine robustness, highlighting the demand for new defense strategies in FL, especially with training on Non-IID datasets.
	
	Our key contributions can be summarized as follows:
	\begin{itemize}
	    \item We survey existing Byzantine attack strategies to compromise FL, as well as Byzantine-robust aggregation schemes that aim to defend against Byzantine attacks.
	    \item Based on an existing clustering-based aggregation scheme, we further propose an enhanced scheme called \ours, by applying an automatical clipping technique to mitigate the effect of amplified local updates.
	    \item We evaluate eight robust aggregation schemes (including the proposed \ours) under five representative Byzantine attack strategies. Our experimental results show that the aggregation schemes sustain high accuracy in some cases, but have limited success in other cases, especially in the presence of Non-IID data.
	\end{itemize}
	
	The rest of this paper is organized as follows: Section~\ref{formulation} first formulates the problem of FL and introduces two optimization algorithms. Then Section~\ref{sec_threat} introduces the threat models evaluated in this paper. Subsequently, representative robust aggregation schemes are presented in Section~\ref{aggs}. Section~\ref{sec_adp} presents an adaptive attack to the proposed aggregation scheme, \ours. Section~\ref{evaluation} presents the experiments for robust aggregation schemes, from which some notable findings are uncovered. Finally, we review related work in Section~\ref{relatedwork} and make some conclusions in Section~\ref{conclusion}.

	\section{Federated Learning}
	
	\label{formulation}
	In this section, we first formulate the optimization problem of FL. Then we introduce two popular algorithms for solving the FL problem, one is the classic distributed SGD optimization algorithm \fedsgd and the other is the famous communication-efficient algorithm \fedavg. 
	
	\subsection{Problem Formulation}
	\label{sec_formulation}
	In FL, multiple clients collaboratively learn a shared global model using their private datasets in a distributed way, assisted by the coordination of a central server. The goal is to find a parameter vector $\bm{w}$ that minimizes the following distributed optimization model:
	\begin{equation}
		\label{obj}
		\min\limits_{\bm{w}} F(\bm{w}) = \min\limits_{\bm{w}}  \frac{1}{K} \sum_{k \in [K]}  F_k(\bm{w}),
	\end{equation}
	where $K$ is the total number of clients, the local objective $F_k(\cdot)$ can be defined as an empirical risk over local data, i.e., $F_k(w) = \frac{1}{n_k}\sum_{j \in [n_k]}\loss(\bw;x_{k,j})$, where $\loss(\cdot;\cdot)$ is a user-specified loss function, $x_{k,j}$ is a training sample and $n_k$ is the size of training dataset owned by client $k$.
	
	A common assumption in FL is that local training datasets can be unbalanced, i.e., clients can have different numbers of training samples~\cite{mcmahan2017communication}. However, in this paper, we assume that data are balanced, i.e., $n_1=n_2=\dots=n_K$ to align with most studies that specifically focus on Byzantine robustness~\cite{chen2017distributed,yin2018byzantine,karimireddy2021learning}. We note that one can get rid of this assumption using the re-scaling trick proposed by Li et al.~\cite{li2019convergence}.

	\subsection{Optimizations of Federated Learning}
	We adopt the two most popular algorithms in Byzantine robust optimization literature to solve Problem~\eqref{obj}, i.e., \fedsgd and \fedavg. 
	
	\begin{algorithm}[t]
		\algdef{SE}[SUBALG]{Indent}{EndIndent}{}{\algorithmicend\ }%
		\algtext*{Indent}
		\algtext*{EndIndent}
		\setlength{\abovedisplayskip}{0pt}
		\setlength{\belowdisplayskip}{0pt}
		\setlength{\abovedisplayshortskip}{0pt}
		\setlength{\belowdisplayshortskip}{0pt}
		\caption{Optimization of Federated Learning}
		\hspace*{\algorithmicindent} \textbf{Input:} $K, T, \eta_t$, $\bw^{0}$
		\label{alg}
		\begin{algorithmic}[1]
			\For{each global round $t \in [T]$}
			\For{each client $k \in [K]$ \textbf{in parallel}}
			\State{$\bm{w}_k^t \leftarrow \bm{w}^t$}
			
			\State{\textbf{Option I} (\fedsgd):}
			\Indent
			\State{Sample mini-batch $\xi$ from local dataset} 
			\State $\update_k^t \leftarrow \grad F_k( \bm{w}_{k} ^ t, \xi) $ \label{grad_cal}
			\EndIndent
			
			\State{\textbf{Option II} (\fedavg):}
			\Indent
			\For{ $E_l$ local rounds, } \label{ag_g_data}
			\State{Sample mini-batch $\xi$ from local dataset} 
			\State{$ {\displaystyle \bm{w}_{k} ^ t \leftarrow  \bm{w}_{k} ^ t - \eta_t \grad F_k( \bm{w}_{k} ^ t, \xi)} $}
			\EndFor 
			\State $\update_k^t \leftarrow \bm{w}_{k}^t - \bm{w}^t$ \label{update_cal}
			\EndIndent
			
			\State{Sends $\Delta_k^t$ back to the server} \label{ag_g_update}
			\EndFor
			\State $\Delta^{t+1} \leftarrow AGG(\{\Delta_k^t\}_{k\in [K]})$
			
			\State{\textbf{Option I} (\fedsgd):}
			\Indent
			\State $\bm{w}^{t+1} \leftarrow \bm{w}^{t} - \eta_t \Delta^{t+1}$
			\EndIndent
			
			\State{\textbf{Option II} (\fedavg):}
			\Indent
			\State $\bm{w}^{t+1} \leftarrow \bm{w}^{t} + \Delta^{t+1}$
			\EndIndent
			\EndFor
			\State \Return $\bw^{T}$ 
		\end{algorithmic}
	\end{algorithm}
	
	\subsubsection{\fedsgd}
	
	Stochastic gradient descent (SGD) can be applied naively to the federated optimization problem~\eqref{obj}~\cite{mcmahan2017communication}. As summarized in Algorithm~\ref{alg} with option I, each client calculates a single mini-batch gradient and uploads it to the server in parallel at each round of training. The server then aggregates the received gradients and updates the model parameters according to the aggregated gradients. Benefiting from mini-batch of stochastic gradient calculation, 
	this approach is computationally efficient, but it still requires a very large number of communication rounds to produce good models~\cite{li2020preserving,mcmahan2017communication}. In this paper, we refer to this algorithm as \fedsgd, also known as \syncsgd in some related work~\cite{baruch2019little,xie2020fall}.
	
	\subsubsection{\fedavg}
	
	A more communication-efficient framework for FL is \fedavg~\cite{mcmahan2017communication}. As summarized in Algorithm~\ref{alg} with Option II, at each round of training, the server broadcasts its global model to each client. In parallel, the clients run multiple steps of SGD on their own loss functions and send the resulting model to the server. The server then updates its global model according to its aggregation rule and broadcasts the resulting global model to each client to enable the next round of training. Multiple rounds of interactions between the server and clients are required to obtain an accurate shared global model.
    
    As one may see, general \fedavg-based algorithms usually randomly select a subset of clients to perform local training while the algorithm we adopt involves full participation of all clients at each round. This is because all of the aggregation schemes considered in this paper are based on an assumption that less than half of the updates for aggregation are malicious on each round. Selecting subsets at random violates this assumption with some probability, as it may select more malicious clients than benign ones by chance. Therefore, full participation is used in this paper.
    
	\subsubsection{Update aggregation}
	In both \fedsgd and \fedavg, the server aggregates the received updates and uses the result of the aggregation to update the global model. A widely-used aggregation scheme is calculating the sample \mean of the uploaded updates, i.e.,
	\begin{equation}
		\Delta^{t+1} \leftarrow \frac{1}{K} \sum_{k \in [K]} \Delta_k^t.
	\end{equation}
	However, \mean is vulnerable to malicious local updates. As the breakdown point of \mean is $1/K$~\cite{lopuhaa1991breakdown}, which means that even if only one of the clients is malicious, the resulting  global model can significantly deviate from the original \mean. In Section~\ref{aggs} we will cover robust aggregations that aim to defend against malicious updates.
	
\begin{figure}[t]
	\centering
	\includegraphics[width=\linewidth]{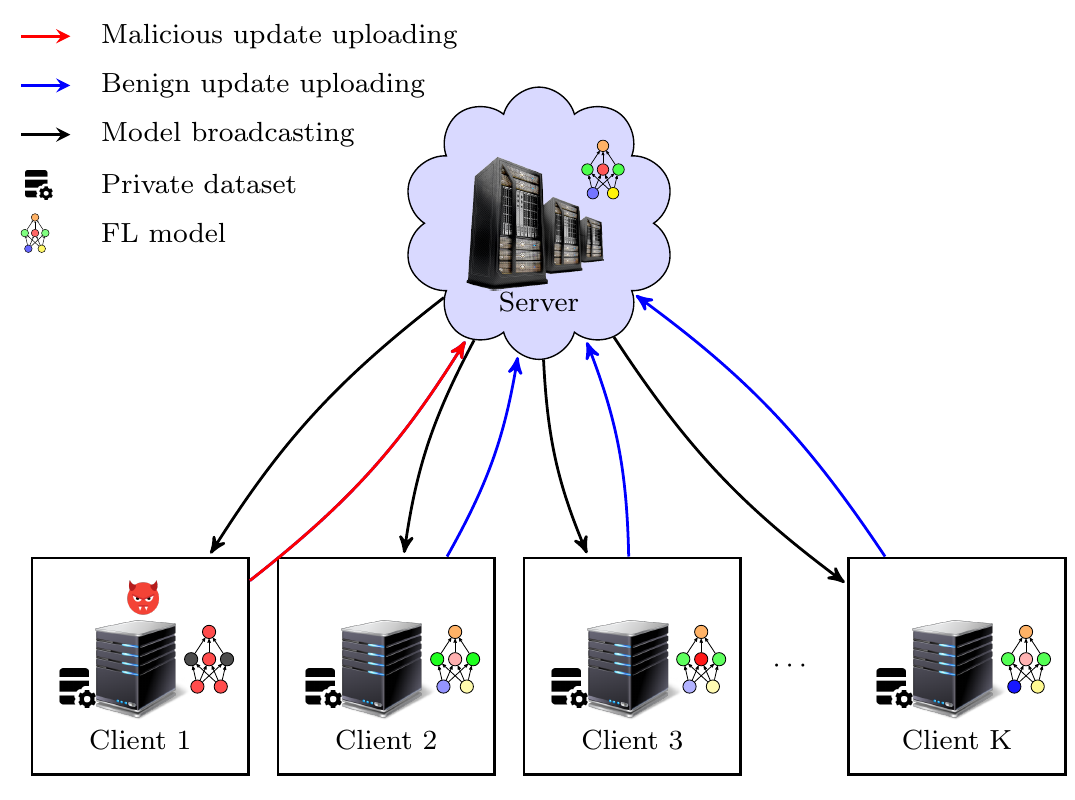}
	\caption{\small Illustration of an FL system with Byzantine clients, where $K$ clients collaboratively train a machine learning model using their local datasets. The coordinate server aggregates local updates and updates the model, then broadcasts the new model to the clients in each communication round. The attackers control some of the clients (e.g., Client 1) and send malicious updates to the server, while the honest clients compute and upload benign updates.}
	\label{arch}
\end{figure}

	\section{Threat Models}
	\label{sec_threat}
	In this section, we describe the five threat models of Byzantine attacks we evaluate in this paper.
	
	In terms of Byzantine attacks, 
	most existing literature for distributed learning and federated learning focuses on convergence prevention~\cite{xie2020fall,fang2020local,li2021byzantine,sattler2020byzantine}. As illustrated in Fig.~\ref{arch}, the attackers (known as Byzantine clients) may upload arbitrary parameters to the server in order to degrade the performance of the global model. Thus, Line~\ref{grad_cal} and \ref{update_cal} of Algorithm~\ref{alg} are replaced by the following:
	\begin{align}
		\label{inner_solution}
		\update_k^t &\leftarrow
		\begin{cases}
			\star            & \text{if $k$-th client is Byzantine,} \\
			\grad F_k( \bm{w}_{k} ^ t, \xi) & \text{if $k$-th client is benign (for \fedsgd),} \\
			\bm{w}_{k}^t - \bm{w}^t & \text{if $k$-th client is benign (for \fedavg),}
		\end{cases}
	\end{align} where $\star$ represents arbitrary values.
	
	In this paper, we follow the assumption that the majority of the clients are benign~\cite{li2021byzantine,karimireddy2021learning}, which means we have $\frac{M}{K} < 0.5$, where $M$ is the number of Byzantine clients. We examine five typical attacks in our threat models.

	\subsection{Noise}
	
	A straightforward attack is to sample some random noise from a distribution (e.g., Gaussian distribution) and add it to the updates before uploading~\cite{li2021byzantine,li2021ditto}. For simplicity sake, the mean and variance of the noise are both $0.1$ in our experiment.
	
	\subsection{A Little is Enough (ALIE)}
	In contrary to the random Noise attack, the attackers may modify the noise carefully to pretend to be benign and fool the aggregation rules. A Little is Enough (ALIE)~\cite{baruch2019little} assumes that the benign updates are expressed by a normal distribution. The attackers therefore immediately take advantage of the high empirical variance between the updates of clients and upload a noise in a range without being detected.
	
	For each coordinate $i \in [d]$, the attackers calculate mean ($\mu_i$) and std ($\delta_i$) over benign updates, and set corrupted updates $\Delta_i$ to values in the range $(\mu_i - z^{max}\delta_i, \mu_i + z^{max}\delta_i)$, where $z^{max}$ ranges from 0 to 1, and is typically obtained from the Cumulative Standard Normal Function~\cite{baruch2019little}.
	
	\subsection{Inner Product Manipulation (IPM)}
	\label{sec_ipm}
	The Inner Product Manipulation (IPM) attack~\cite{xie2020fall} seeks for the negative inner product between the true mean of the updates and the output of the aggregation schemes so that at least the loss will not descend. Assuming that the attackers know the mean of benign updates, a specific way to perform an IPM attack is
	\begin{equation}
		\update_1^t = \dots = \Delta_M^t = - \frac{\epsilon}{K - M}\sum_{i = M+1}^K \Delta_i^t,
	\end{equation} where we assume that the first $M$ clients are malicious, $\epsilon$ is a positive coefficient controlling the magnitude of malicious updates. Then the \mean becomes
	\begin{equation}
		\frac{1}{K} \sum_{k \in [K]} \Delta_k^t = \frac{K - M(1+\epsilon)}{K(K - M)}\sum_{i = M+1}^K \Delta_i^t.
	\end{equation} 
	
	Note that when $\epsilon < \frac{K}{M} - 1$, IPM does not change the direction of the average over benign updates but only decreases its magnitude, because we have
	$$
	\frac{K - M(1+\epsilon)}{K(K - M)} > 0,
	$$ the optimization thus can still converge using \mean as an aggregation scheme. However, as we will show in Section~\ref{evaluation}, such an attack can circumvent the defense of several aggregation schemes and inverse the direction of updates, which heavily damages the global model. On the contrary, when $\epsilon > \frac{K}{M} - 1$, the sign of \mean is reversed, indicating that the loss will increase if the model is updated using the \mean. In our experiment, we examine both cases by letting $\epsilon=0.5$ and $\epsilon=100$, respectively.
	
	\subsection{Sign Flipping (SF)}
	Different from IPM, the Sign Flipping (SF) attackers do not need to know the updates from other clients and simply flip the signs of the gradient~\cite{li2019rsa,karimireddy2021learning}, which means that the attackers strive to maximize the loss via gradient ascent instead of gradient descent. Specifically, in \fedsgd, the clients upload the negative gradients; in \fedavg, the flipping is applied at every local updating step. 
	
	\subsection{Label Flipping (LF)}
	The aforementioned attacks assume that the attackers have full access to the training process so that they can modify the updates immediately. However, full access may be limited as the training APIs are not always open. Correspondingly, the attackers can also change the training dataset instead of the update parameters~\cite{jere2020taxonomy}. The Label Flipping (LF) attack simply flips the label of each training sample~\cite{fang2020local}. Specifically, a label $l$ is flipped as $L-l-1$, where $L$ is the number of classes in the classification problem and $l=0,1,...,L-1$.
	
	\input{defense}



	

	\section{Evaluation}
	\label{evaluation}

	In this section, we design experiments to evaluate the aforementioned robust aggregation schemes under different attacks and show the experimental results. 
	

	\subsection{Experimental Setup}
	We simulate an FL system with the server and 20 clients, targeting at image classification tasks on CIFAR-10~\cite{Krizhevsky09learningmultiple} and MNIST~\cite{lecun2010mnist} datasets with both IID and Non-IID partitions, where five of the clients are Byzantine clients by default. For the IID partitioning, we randomly split the training set into 20 subsets and allocate them to the 20 clients. For the Non-IID partition, we follow prior work~\cite{lin2020ensemble,arflli} and model the Non-IID data distributions with a Dirichlet distribution $\bm{p}_l \sim Dir_K(\alpha)$. Then we allocate a $\bm{p}_{l,k}$ proportion of the training sample of class $l$ to client $k$, in which a smaller $\alpha$ indicates a stronger Non-IID data partition. We let $\alpha=0.1$ for all Non-IID settings. Fig.~\ref{fig_noniid} visualizes the resulting statistical heterogeneity of labels on CIFAR-10. Such a partition is strongly Non-IID as one can see that some of the classes are completely missing for each client.
	
	For CIFAR-10, we choose a lightweight Compact Convolutional Transformers (CCT) network~\cite{hassani2021escaping}, as such a small yet effective model has more potential to overcome the on-board resource limitation of FL devices~\cite{li2020federated}. For MNIST, we train a simple two-layer perceptron using ReLu as activation function. We train the model for 6000 and 600 communication rounds for \fedsgd and \fedavg, respectively. By default, we set the batch size to 128 and 64 for MNIST and CIFAR-10, respectively. As suggested by~Karimireddy et al.~\cite{karimireddy2021learning}, we decay the learning rate during training to improve convergence, i.e., for \fedsgd, we let 
	\begin{align}
		\eta_t &\leftarrow
		\begin{cases}
			0.1           & \text{if $t \le 2000$,} \\
			0.05 & \text{if $2000 < t \le 5000$,} \\
			0.025 & \text{else};
		\end{cases}
	\end{align}
	for \fedavg, we let
	\begin{align}
		\eta_t &\leftarrow
		\begin{cases}
			0.1           & \text{if $t \le 200$,} \\
			0.05 & \text{if $200 < t \le 500$,} \\
			0.025 & \text{else}.
		\end{cases}
	\end{align} For \fedavg, we apply 50 SGD steps before uploading the updates to the server (i.e., $E_l = 50$).

	\subsection{Impact on the \mean Scheme}
	We first demonstrate the impact of attacks on conventional \fedsgd and \fedavg using the \mean scheme to aggregate updates by plotting test accuracy versus the number of communication rounds in Fig.~\ref{fig_mean_agg}. At first glance, when the datasets are IID, \fedavg takes fewer communication rounds than \fedsgd to converge when there is no attack, benefiting from multiple steps of local training. However, when the datasets are Non-IID, the performance of \fedavg significantly decreases while \fedsgd still maintains relatively high accuracy. We note that this is a well-known drawback of \fedavg 
	\cite{karimireddy2020scaffold,deng2020adaptive}.
	
	Recall that some of the attacks (e.g., IPM with small $\epsilon$~\cite{xie2020fall}) are particularly designed to break the line of robust defense, which means that they may not bring much damage to the \mean scheme. For instance, Fig.~\ref{fig_mean_agg} shows that IPM ($\epsilon = 0.5$) produces less damage to \mean compared to the other attacks, as the malicious updates do not change the direction of the average update but only decrease its magnitude (see Section~\ref{sec_ipm}). Furthermore, Noise and IPM ($\epsilon=100$) eventually damage the models under the four settings by decreasing the test accuracy to around $10\%$ (no better than random guess). This is because they both make large changes to the updates, and \mean could be easily biased by large changes~\cite{li2021byzantine}.

    Unsurprisingly, ALIE attack is ineffective to MNIST dataset, while it makes a large impact on CIFAR-10 with \fedsgd. This is because the amount of noise for ALIE is determined by the empirical variance of benign updates. As it is already known that MLP on MNIST usually has a lower variance than CNN on CIFAR-10~\cite{faghri2020study}.
    
	\newcommand \rotv{90}
	\newcommand{\rotboxv}[1]{\rotatebox[origin=c]{\rotv}{#1}}
	\begin{table*}[ht]
	\begin{center}
            \setlength\extrarowheight{2pt}
		\begin{adjustbox}{width=2.05\columnwidth,center}
		\begin{tabular}{|c|c|c||c|c|c|c|c|c|c||c|c|c|c|c|c|c|}
			\hline
			\rotbox{ \ Dataset \ } & \rotbox{ \ Algorithm \ } & Defense & \multicolumn{7}{c||}{IID data} & \multicolumn{7}{c|}{Non-IID data} \\
			 \hline
			 & & & \multicolumn{1}{c|}{ \makecell{No \\ Attack} }  & \makecell{IPM \\ { \scriptsize ($\epsilon$=0.5)}} & \makecell{IPM \\ {\scriptsize ($\epsilon$=100)}} & SF & LF & ALIE & Noise & \multicolumn{1}{c|}{ \makecell{No \\ Attack} } &  \makecell{IPM \\{\scriptsize ($\epsilon$=0.5)}} & \makecell{IPM \\ { \scriptsize($\epsilon$=100)}} & SF & LF & ALIE & Noise \\
			 \hline
            \multirow{9}{*}{\rotboxv{MNIST}} & \multirow{9}{*}{\rotboxv{\fedsgd}} & \mean & \textbf{97.13} & 96.16 & \transparent{0.3}10.32 & 95.19 & 96.70 & 96.98 & \transparent{0.3}11.35 & 96.70 & \textbf{95.35} & \transparent{0.3}11.35 & 18.18 & 79.45 & \textbf{96.14} & \transparent{0.3}9.80 \\
			 &  & \krum & 96.04 & 88.64 & 96.15 & \textbf{95.51} & 96.15 & 95.96 & 96.15 & 78.57 & 52.28 & 81.54 & 66.37 & 82.07 & 93.82 & 81.54 \\
			 &  & \gm & 95.97 & 77.12 & 95.99 & 95.47 & 96.12 & 95.96 & \textbf{96.19} & 57.59 & 47.57 & 55.94 & 49.07 & 58.16 & 93.74 & 50.08 \\
			 &  & \autogm & 95.97 & 77.12 & 96.15 & 95.47 & 96.08 & 95.96 & 96.19 & 58.51 & 46.86 & 55.75 & 48.76 & 59.99 & 93.74 & 50.23 \\
			 &  & \median & 96.07 & 90.65 & 89.10 & 94.78 & 95.27 & 96.11 & 94.68 & 80.10 & 15.62 & 19.52 & 52.39 & 72.12 & 92.90 & 79.96 \\
			 &  & \tm & 97.13 & 93.92 & 88.97 & 95.02 & 95.24 & 96.27 & 93.63 & 96.70 & 21.52 & \transparent{0.3}12.60 & 76.85 & 77.19 & 92.53 & \textbf{89.43} \\
			 &  & \cc & 97.13 & 96.16 & \transparent{0.3}0.79 & 95.20 & 96.70 & 96.98 & \transparent{0.3}10.09 & 96.69 & 95.35 & \transparent{0.3}11.35 & 82.70 & 79.46 & 96.14 & \transparent{0.3}9.58 \\
			 &  & \clustering & 97.11 & 96.07 & \transparent{0.3}10.32 & 95.26 & 96.72 & \textbf{97.00} & \transparent{0.3}11.35 & \textbf{96.79} & 95.31 & \transparent{0.3}11.35 & \transparent{0.3}9.74 & 80.95 & 95.88 & \transparent{0.3}10.32 \\
			 \hline 
			 &  & \makecell{\footnotesize \ours \\ (ours)} & 97.04 & \textbf{96.90} & \textbf{96.98} & 95.50 & \textbf{96.80} & 96.93 & 92.71 & 95.91 & 85.19 & \textbf{84.82} & \textbf{88.71} & \textbf{85.28} & 95.56 & \transparent{0.3}10.07 \\
			\hline
			\hline
            \multirow{9}{*}{\rotboxv{MNIST}} & \multirow{9}{*}{\rotboxv{\fedavg}} &  \mean & 98.13 & 97.83 & \transparent{0.3}11.35 & \transparent{0.3}11.35 & 97.66 & 97.87 & \transparent{0.3}11.35 & 96.87 & \textbf{96.36} & \transparent{0.3}11.35 & \transparent{0.3}8.92 & 87.02 & \textbf{96.91} & \transparent{0.3}9.80 \\
			 &  & \krum & 95.23 & 93.61 & 95.35 & 95.35 & 95.35 & 97.86 & 95.35 & 73.06 & 42.03 & 74.81 & 74.81 & 58.06 & 96.44 & 74.81 \\
			 &  & \gm & 95.24 & 93.17 & 93.77 & 95.22 & 95.13 & 97.86 & 95.36 & 46.64 & 55.75 & 26.09 & 48.90 & 48.15 & 96.44 & 46.68 \\
			 &  & \autogm & 95.24 & 85.45 & 95.21 & 95.17 & 95.13 & 97.86 & 95.48 & 46.69 & 55.75 & 48.23 & 48.94 & 49.09 & 96.44 & 46.68 \\
			 &  & \median & 98.08 & 96.72 & 96.09 & 97.78 & 97.80 & 97.82 & \textbf{98.02} & 89.85 & 58.06 & 32.16 & 86.97 & 91.09 & 96.14 & 92.43 \\
			 &  & \tm & 98.13 & 97.64 & 95.73 & \textbf{97.83} & 97.78 & 97.85 & 98.01 & 96.87 & 61.93 & 55.51 & 91.77 & 93.03 & 95.26 & \textbf{94.98} \\
			 &  & \cc & 98.10 & 97.84 & \transparent{0.3}9.82 & \transparent{0.3}11.35 & 97.67 & \textbf{97.89} & \transparent{0.3}11.35 & \textbf{96.88} & 96.35 & \transparent{0.3}11.35 & \transparent{0.3}9.74 & 87.00 & 96.91 & \transparent{0.3}9.80 \\
			 &  & \clustering & 98.11 & 97.87 & \transparent{0.3}11.35 & \transparent{0.3}9.74 & 97.81 & 97.84 & \transparent{0.3}10.28 & 96.85 & 96.04 & \transparent{0.3}11.35 & \transparent{0.3}9.58 & 87.70 & 96.86 & \transparent{0.3}9.80 \\
			 \hline
			 &  & \makecell{\footnotesize \ours \\ (ours)} & \textbf{98.15} & \textbf{97.93} & \textbf{97.90} &  \textbf{97.83} & \textbf{97.90} & 97.86 & 97.99 & 96.48 & 96.22 & \textbf{81.08} & \textbf{93.80} & \textbf{94.74} & 96.83 & 93.56 \\
			\hline
			\hline
			
			\hline
			 \multirow{9}{*}{\rotboxv{CIFAR-10}} & \multirow{9}{*}{\rotboxv{\fedsgd}}& \mean & \textbf{82.35} & \textbf{81.22} & \transparent{0.3}10.00 & 57.62 & 75.86 & \transparent{0.3}10.00 & \transparent{0.3}10.00 & \textbf{81.15} & 79.44 & \transparent{0.3}10.00 & 26.48 & \textbf{67.98} & \transparent{0.3}10.00 & \transparent{0.3}10.00 \\
			 &  & \krum & 67.47 & 58.17 & 50.60 & 65.24 & 67.90 & \transparent{0.3}10.00 & 60.26 & \transparent{0.3}10.00 & \transparent{0.3}10.00 & \transparent{0.3}10.00 & 17.54 & \transparent{0.3}10.00 & \transparent{0.3}10.00 & \transparent{0.3}10.00 \\
			 &  & \gm & 63.25 & 59.40 & 36.79 & 65.29 & 66.52 & 26.21 & 22.55 & 17.09 & \transparent{0.3}10.00 & 16.21 & 18.98 & \transparent{0.3}10.00 & \transparent{0.3}10.00 & \transparent{0.3}10.00 \\
			 &  & \autogm & 66.16 & 42.70 & 66.69 & 65.16 & 66.55 & \transparent{0.3} 10.00 & 48.06 & \transparent{0.3}10.00 & \transparent{0.3}10.00 & \transparent{0.3}10.00 & 18.44 & 18.18 & \transparent{0.3}10.00 & 18.97 \\
			 &  & \median & 78.78 & 44.42 & 33.31 & 68.76 & 59.60 & \transparent{0.3}10.00 & \transparent{0.3}10.00 & \transparent{0.3}10.00 & \transparent{0.3}10.00 & \transparent{0.3}10.00 & \transparent{0.3}10.00 & \transparent{0.3}10.00 & \transparent{0.3}10.00 & \transparent{0.3}10.08 \\
			 &  & \tm & 82.06 & 51.30 & 34.97 & 68.49 & 54.63 & \transparent{0.3}10.00 & 16.33 & 80.98 & \transparent{0.3}10.00 & \transparent{0.3}10.00 & \transparent{0.3}10.00 & \transparent{0.3}10.00 & \transparent{0.3}10.00 & \transparent{0.3}10.00 \\
			 &  & \cc & 82.21 & 81.20 & \transparent{0.3}8.76 & 15.76 & 23.08 & 38.81 & \transparent{0.3}10.00 & 53.44 & \textbf{80.23} & \transparent{0.3}10.00 & \transparent{0.3}10.00 & 67.79 & \transparent{0.3}10.00 & \transparent{0.3}10.00 \\
			 &  & \clustering & 81.80 & 79.64 & \transparent{0.3}10.00 & 61.89 & 76.37 & \textbf{77.48} & \transparent{0.3}10.00 & 79.37 & 77.47 & \transparent{0.3}10.00 & 24.77 & 66.35 & \transparent{0.3}10.00 & \transparent{0.3}10.00 \\
			  \hline
			 &  & \makecell{\footnotesize \ours \\ (ours)} & 81.58 & 80.34 & \textbf{81.32} & \textbf{77.94} & \textbf{79.18} & \transparent{0.3}10.00 & \textbf{73.89} & 76.83 & 69.53 & \textbf{20.06} & \textbf{52.37} & 64.48 & \textbf{16.84} & \textbf{20.28} \\
			 \hline
			 \hline
    			 \multirow{9}{*}{\rotboxv{CIFAR-10}} & \multirow{9}{*}{\rotboxv{\fedavg}}& \mean & 80.19 & 76.18 & \transparent{0.3}10.00 & \transparent{0.3}10.00 & 69.45 & 78.10 & \transparent{0.3}10.00 & 63.35 & 56.89 & \transparent{0.3}10.00 & \transparent{0.3}10.00 & 51.60 & 58.89 & \transparent{0.3}10.00 \\
			 &  & \krum & 72.04 & 54.53 & 72.32 & 70.41 & 72.22 & 75.55 & 71.92 & 27.09 & \transparent{0.3}10.00 & 21.47 & 23.26 & 26.98 & 47.36 & 21.78 \\
			 &  & \gm & 72.93 & 49.76 & 69.80 & \transparent{0.3}10.00 & 73.72 & 75.97 & 71.92 & 27.95 & \transparent{0.3}10.00 & \transparent{0.3}10.00 & \transparent{0.3}10.00 & 24.19 & 48.45 & 26.38 \\
			 &  & \autogm & 73.13 & 44.64 & 69.11 & 71.40 & 72.88 & 75.39 & 71.13 & 30.12 & \transparent{0.3}10.00 & 21.68 & 26.21 & 20.30 & 48.92 & 29.17 \\
			 &  & \median & 80.74 & 66.39 & 65.01 & 73.86 & \textbf{77.94} & 76.48 & 75.43 & 57.62 & 20.60 & 16.96 & \textbf{48.60} & 45.22 & 49.49 & 42.58 \\
			 &  & \tm & \textbf{80.76} & 73.42 & 61.13 & 66.52 & 76.42 & 77.66 & 73.08 & 62.98 & 30.92 & 18.61 & 47.95 & 47.92 & 53.03 & 39.67 \\
			 &  & \cc & 80.01 & 77.77 & \transparent{0.3}10.00 & \transparent{0.3}10.00 & 69.80 & \textbf{78.53} & 24.71 & \textbf{63.39} & 56.97 & \transparent{0.3}10.00 & \transparent{0.3}10.00 & \textbf{52.41} & \textbf{59.29} & 17.24 \\
			 &  & \clustering & 80.15 & 77.74 & \transparent{0.3}10.00 & \transparent{0.3}10.00 & 75.61 & 78.12 & \transparent{0.3}10.00 & 61.45 & \textbf{57.37} & \transparent{0.3}10.00 & \transparent{0.3}10.23 & 49.43 & 58.66 & \transparent{0.3}10.00 \\
			 \hline
			 &  & \makecell{\footnotesize \ours \\ (ours)} & 79.71 & \textbf{78.80} & \textbf{78.44} & \textbf{78.01} & 63.07 & 77.73 & \textbf{78.82} & 62.52 & 53.31 & \textbf{31.24} & 24.31 & 50.77 & 56.51 & \textbf{48.55} \\
			 \hline
		\end{tabular}
		\end{adjustbox}
	\end{center}
	\caption{Comparing state-of-the-art robust aggregation schemes and our \ours under various attacks, where 25\% of the clients are malicious. A semi-transparent value indicates that the corresponding accuracy is lower than 15\% (not much better than random guessing). We bold the numbers with the highest accuracy. When integrated with robust aggregation schemes, \fedavg is more robust than \fedsgd. On the other hand, all the attacks become more effective in Non-IID scenarios.}
	\label{overall_result}
    \end{table*}

	\subsection{Impact on Robust Aggregation Schemes}
	Table~\ref{overall_result} shows the overall comparison of the robust aggregation schemes in \fedsgd and \fedavg with respect to test accuracy on both IID and Non-IID partitioned datasets.
	
	First of all, with the absence of malicious clients, \krum, \gm, \autogm, and \median achieve relatively lower accuracy compared to other schemes. Especially, such degradation is more significant with Non-IID data. For instance, the accuracy almost reduces to random guessing, 10\%, for CIFAR-10 with \fedsgd, while \mean, \tm, \clustering, and \ours sustain the accuracy at around 80\%. This suggests that \krum, \gm, \autogm, and \median should be used with caution for highly Non-IID data.
	
	Euclidean-based schemes (i.e., \krum, \gm, and \autogm) reach lower accuracy compared to the other schemes in the complete absence of attackers, especially when the datasets are Non-IID. Surprisingly, their performance of \fedsgd with Non-IID CIFAR-10 data is not much better than random guessing. This might be because they all tend to select a single update that is closest to all or part of the others measured by Euclidean distance, which however is a poor estimation of the overall tendency in data heterogeneity situations. Furthermore, they all show similar robustness in both \fedsgd and \fedavg. E.g., with IID data partition, they all handle Noise and IPM ($\epsilon = 100$) well while somehow struggling with the other attacks. This is not surprising as those Euclidean-based schemes are essentially designed to defend against large changes of updates. On the other hand, small-scale attacks such as IPM ($\epsilon = 0.5$) challenges them as the malicious updates are close to the benign ones when measured by Euclidean distance.
	
	Two other classic robust aggregation schemes, \median, and \tm show similar robustness in most cases. For instance, when tested with IID MNIST, they successfully defend against most attacks with insignificant accuracy degradation. However, for Non-IID MNIST, both suffer from IPM attacks. A similar phenomenon occurs with CIFAR-10.
	
    \cc and \clustering both fail in all IPM ($\epsilon=100$) and Noise attacks, because these attacks are of large magnitude. In these cases, \ours significantly enhances the robustness of \clustering, benefiting from the adaptive clipping mechanism. 
	In the experiments, our proposed \ours successfully defends against more types of attacks than the other schemes, and achieves the highest test accuracy. However, it is defeated by ALIE when we train the model using \fedsgd with CIFAR-10. Similar to other schemes, \ours is also less robust in Non-IID data scenarios. We note that Non-IIDness is a well-known challenge to FL especially when it comes to robustness~\cite{arflli}, as it becomes hard to induce a consensus model for the benign clients if their data distributions are significantly different. Thus the malicious clients can damage the global model more easily.
	
	
	Another important observation from our experiments is that ALIE makes almost no accuracy degradation when we train models using MNIST dataset, while it successfully circumvents almost all aggregation schemes of \fedsgd with CIFAR-10 and keeps the test accuracy below 20\%. The models suffer much less damage from ALIE when trained with \fedavg. Moreover, even the \mean scheme could handle it well. We thus infer that multiple steps of SGD updates may result in lower variance compared to single-step updates when the datasets are IID.

    \begin{figure*}[ht]
		\centering
		\includegraphics[width=\linewidth]{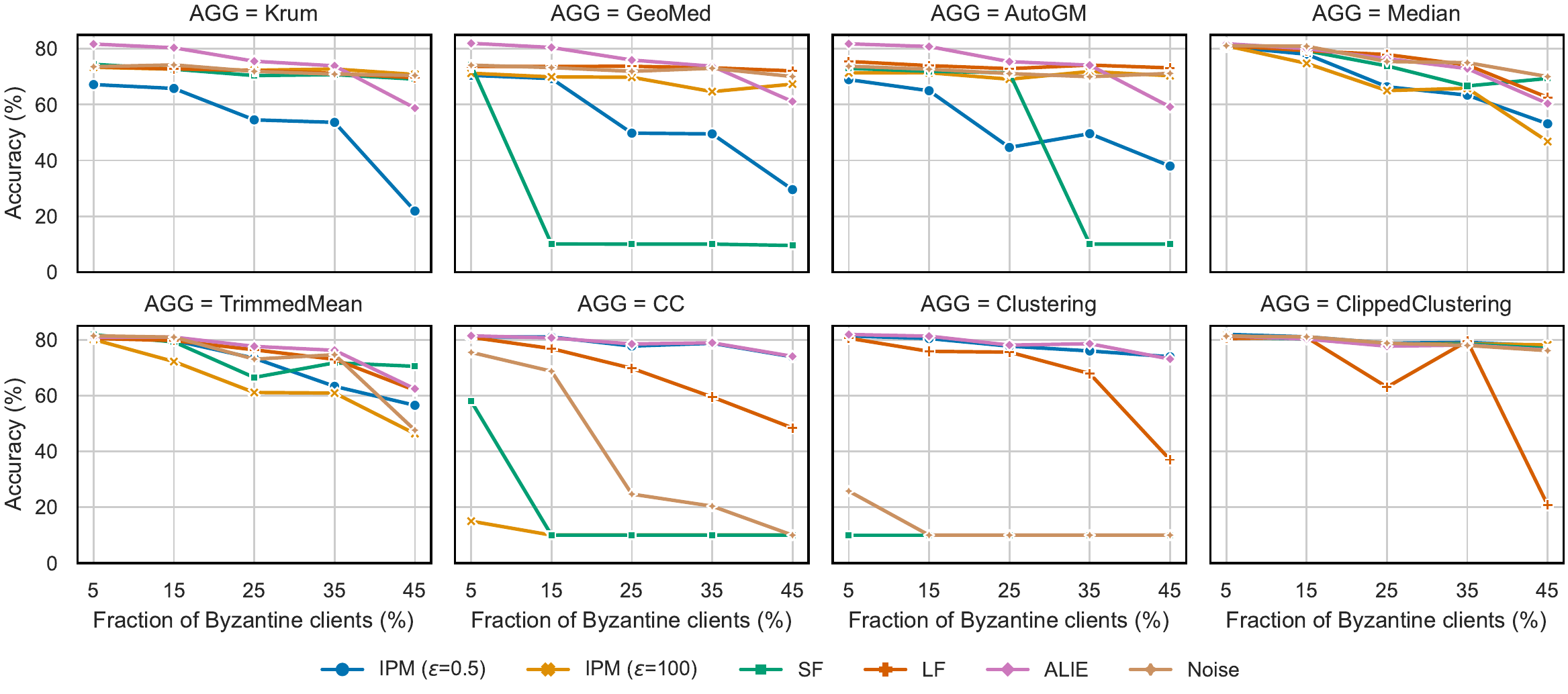}
		\caption{Impact of the fraction of Byzantine clients on test accuracy for CIFAR-10 with \fedavg. \median, \tm, and \ours show high tolerance to most attacks.}
		\label{fig_fraction}
	\end{figure*}
	
	\begin{figure*}[ht]
		\centering
		\includegraphics[width=\linewidth]{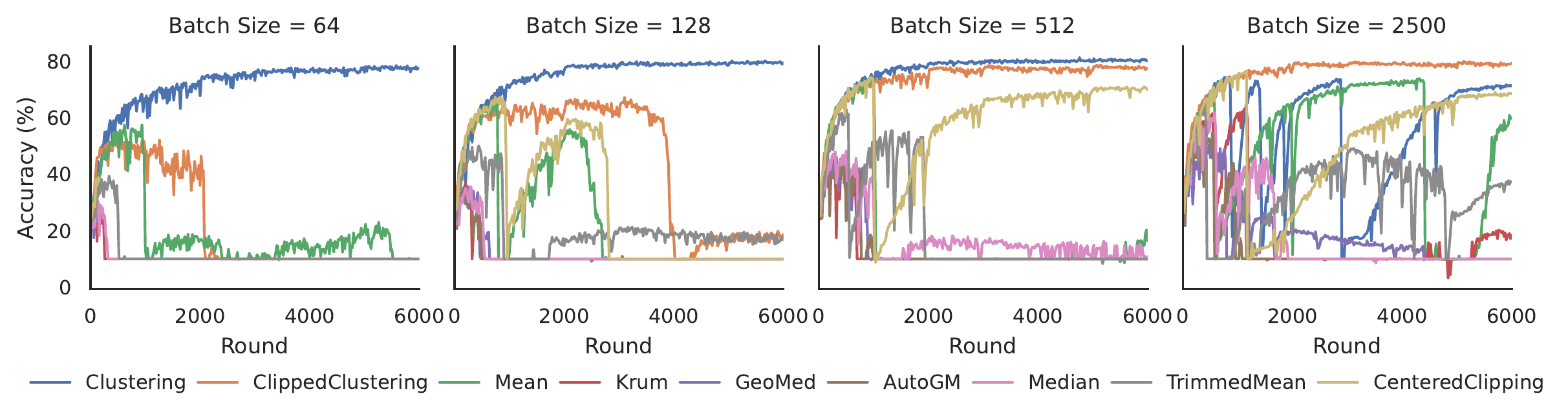}
		\caption{Impact of batch size on the performance of \fedsgd with IID partition under ALIE attack. The aggregation schemes (including \mean) become more robust as the batch size increases.}
		\label{fig_batch_size}
	\end{figure*}
		
	\begin{figure*}[h!]
		\begin{subfigure}[b]{\linewidth}
			\centering
			\includegraphics[width  =\linewidth]{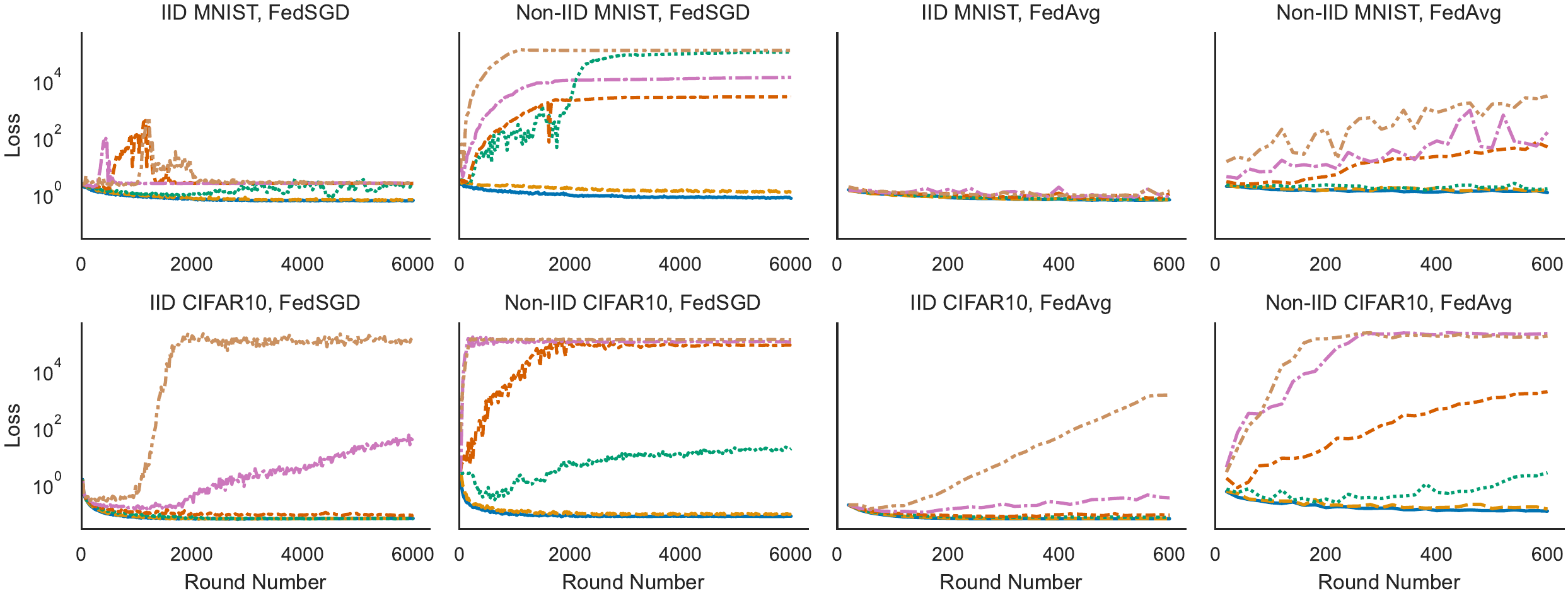}
		\end{subfigure}
		
		\begin{subfigure}[b]{\linewidth}
			\centering
			\includegraphics[width=\linewidth]{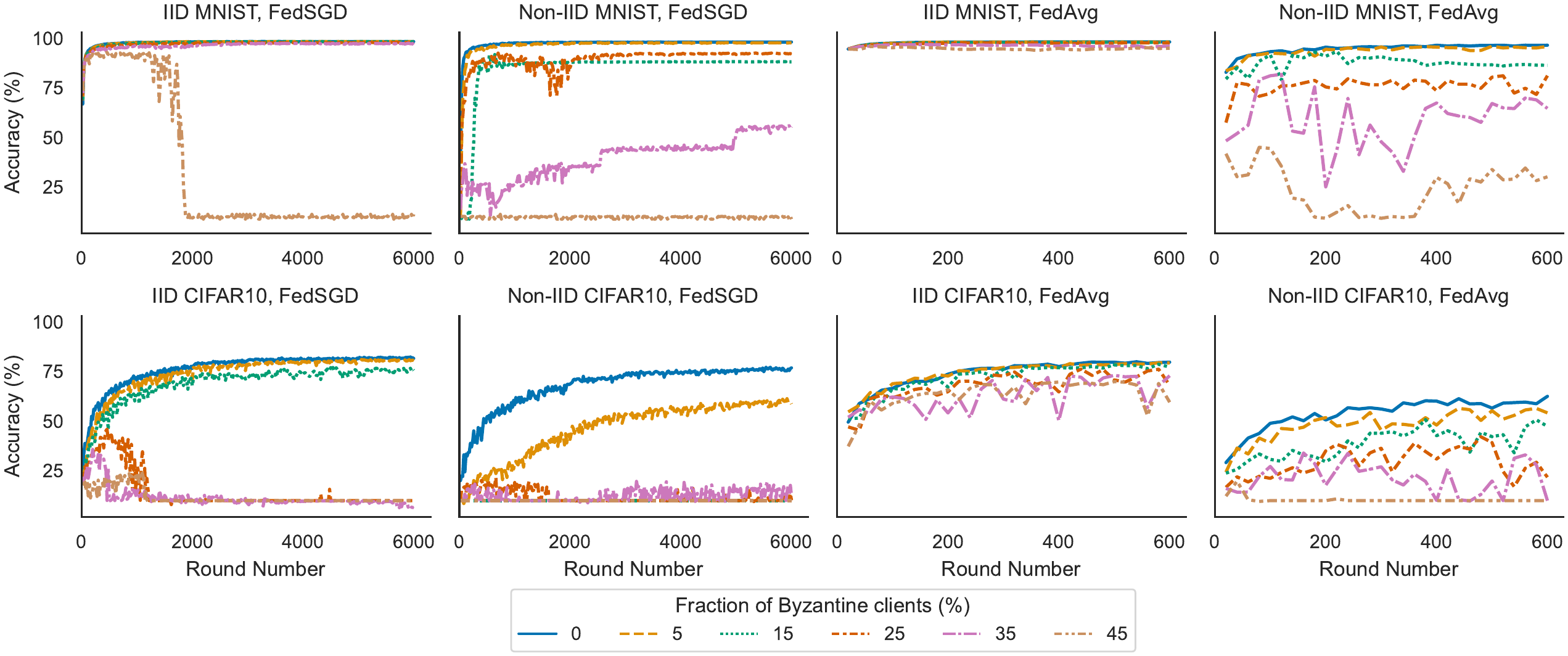}
		\end{subfigure}
		\caption{Adaptive attack to \ours scheme. The top two rows show the test loss and the bottom rows show the test accuracy. We clamp the loss into the range $[0, 10^5]$. The attack is particularly effective when the local datasets are Non-IID.}
		\label{fig_adaptive}
	\end{figure*}	
	
	\begin{figure}[ht]
		\centering
		\includegraphics[width=\linewidth]{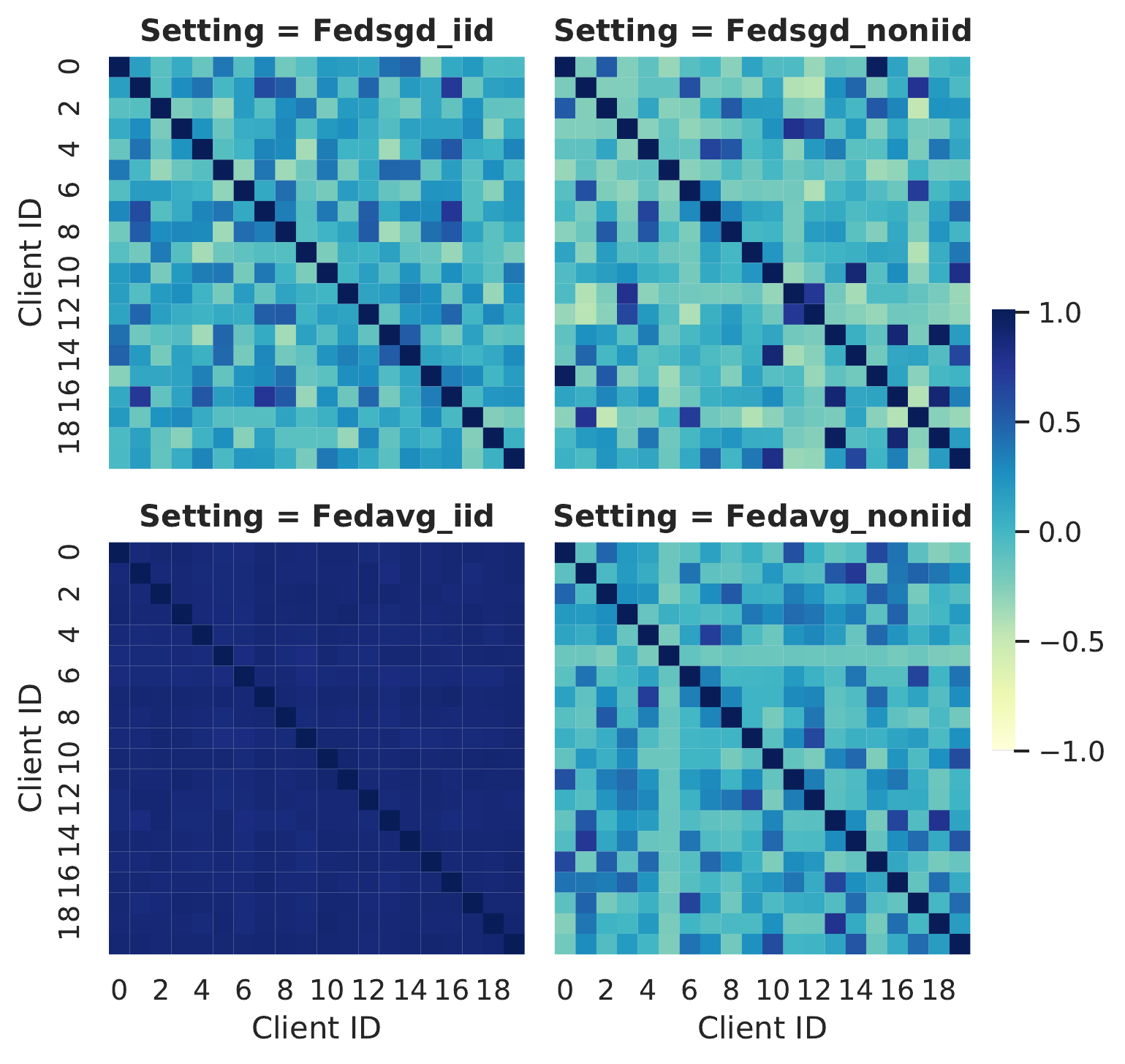}
		\caption{Pairwise cosine similarities between local updates in different scenarios, where a similarly value 1 represents identical update directions, and -1 indicates opposite directions. \fedavg with IID data results in the most similar directions.
		}
		\label{fig_cosine}
	\end{figure}


	\subsection{Impact of Fraction of Malicious Clients}

	To study the impact of the number of malicious clients, we perform an experiment using \fedavg with IID partitioned CIFAR-10 under six types of attacks. The results are shown in Fig.~\ref{fig_fraction}. Overall, the attacks make varying degrees of impact on the aggregation schemes as the fraction of Byzantine clients increases. Interestingly, the tendency of \krum, \gm, and \autogm are similar, except for the impact of SL attack. Particularly, these schemes all suffer from even a small fraction of IPM ($\epsilon=0.5$) attackers, i.e., the accuracy is reduced by 10\% with the presence of a single Byzantine client (when 5\% of the clients are malicious). 
 
    \cc and \clustering are both vulnerable to Noise, SF and IPM ($\epsilon=100$) attacks even with the presence of only $5\%$ Byzantine clients. \median, \tm, and \ours are more robust to attacks especially when the fraction is low. Surprisingly, \ours tends to be affected by LF attacks as the fraction increases. Note that Fig.~\ref{fig_fraction} only shows the performance of \fedavg with IID data. When the local datasets are Non-IID, all schemes show considerably less tolerance to Byzantine attacks. 
		
	\subsection{Impact of Batch Size}
	In the previous experiments, the batch size per client is relatively small (i.e., 128 and 64 for MNIST and CIFAR-10, respectively), which leads to a large variance among benign updates, thus making the attacks more challenging~\cite{karimireddy2021learning}. Taking CIFAR-10 as an example, we investigate the effect of batch size on the robustness of aggregation schemes by varying batch size in $\{64, 128, 512, 2500\}$. 
    
    Fig.~\ref{fig_batch_size} shows the performance of \fedsgd with IID partition under ALIE attacks. \mean, \ours, \autogm, \tm and \cc become more robust as the batch size increases. This is because the variance tends to be lower as we increase the batch size. Particularly, when the batch size is 2500, the clients use all their training data for each step of training, stochastic gradient then becomes population gradient (the optimization becomes full-batch gradient descent). However, except \ours, all the other aggregation schemes still fail to achieve acceptable test accuracy. Although a large batch size is more favorable for robustness purposes as indicated by this experiment, it is not the desired solution as it brings a large computation burden to local training.
	
	\subsection{Impact of Adaptive Attack}
 	We examine \ours with the adaptive attack described in Section~\ref{sec_adp}.  The result is shown in Fig.~\ref{fig_adaptive}. For the sake of visualization, we clamp the loss into the range $[0, 10^5]$. For \fedsgd with IID data (shown in the first column), \ours tolerates up to 15\% clients attacked with little performance degradation. When more than 25\% of clients are malicious, the loss curves fluctuate or even tend to increase. For \fedsgd with Non-IID data (shown in the second column), the models diverge as long as there are 15\% clients are malicious.
	
	As for \fedavg with IID data, \ours shows higher tolerance than \fedsgd. Specifically, the model for IID MNIST successfully converges with all the different fractions of Byzantine clients. However, as shown by the fourth column of Fig.~\ref{fig_adaptive}, it is still much less robust when it comes to Non-IID data, where the model does not tend to converge when more than 15\% of clients are malicious.
	
	Noticeably, we observe that preventing convergence is not always sufficient to degrade the accuracy. For example, for \fedsgd on Non-IID MNIST with 15\% clients are malicious, the loss increases to the upper bound ($10^5$) after 4000 rounds, the model, however, still remains 92\% accuracy. This is because the top-1 accuracy (as we use in this paper) only takes into account the output with the highest probability. In this case, the attackers significantly change the output distribution but fail to change the index of the highest output.
	
	\subsection{Pairwise Cosine Similarities}
	Recall that the aggregation schemes show different robustness in \fedsgd and \fedavg with respective to IID and Non-IID data partitions. To further investigate this issue, we compare the pairwise cosine similarities of all benign local updates without attacks. We note that cosine similarly reflects the angle between two vectors, i.e., a higher value indicates a smaller angle.  As visualized in Fig.~\ref{fig_cosine}, the pairwise cosine similarities of updates from \fedsgd vary widely, no matter if the local datasets are IID or not. Considerable pairs of clients even show negative similarities. Such a stochastic property may confuse the robust aggregation schemes and make it more challenging to detect malicious updates. The updates from \fedavg with IID data show the highest pairwise similarities, which means that their update directions are almost identical. Benefiting from this, clustering-based schemes can group benign updates together and exclude malicious updates. However, they become less similar when the local datasets are Non-IID. 

	\input{related_work.tex}

	\section{Conclusions}
	\label{conclusion}
	In this paper, we provided an experimental study of Byzantine robust aggregation schemes for FL. In particular, we survey existing Byzantine attacks and defense strategies in the FL literature. We also propose a novel scheme, \ours, which enhances the robustness of the clustering-based scheme by automatically clipping the updates to mitigate the effect of amplified malicious updates. We then evaluate eight robust aggregation schemes under five representative Byzantine attack strategies. Our experimental results show that all those aggregation schemes achieve limited robustness in the presence of Byzantine attacks. In the future, it would be interesting to carry out a theoretical analysis to guarantee the robustness of \ours. Furthermore, we plan to improve the robustness of FL from more perspectives, e.g., low variance algorithms, and robust learning rates.

    \section*{Status and Publication Information}
    This paper has been accepted for publication in IEEE Transactions on Big Data, and the final version is now available at \href{https://doi.org/10.1109/TBDATA.2023.3237397}{\url{https://doi.org/10.1109/TBDATA.2023.3237397}}

	\ifCLASSOPTIONcompsoc
	\else
	\fi

	
	\bibliographystyle{IEEEtran}
	\bibliography{reference}
	
		\begin{IEEEbiography}[{\includegraphics[width=1in,height=1.25in, clip,keepaspectratio]{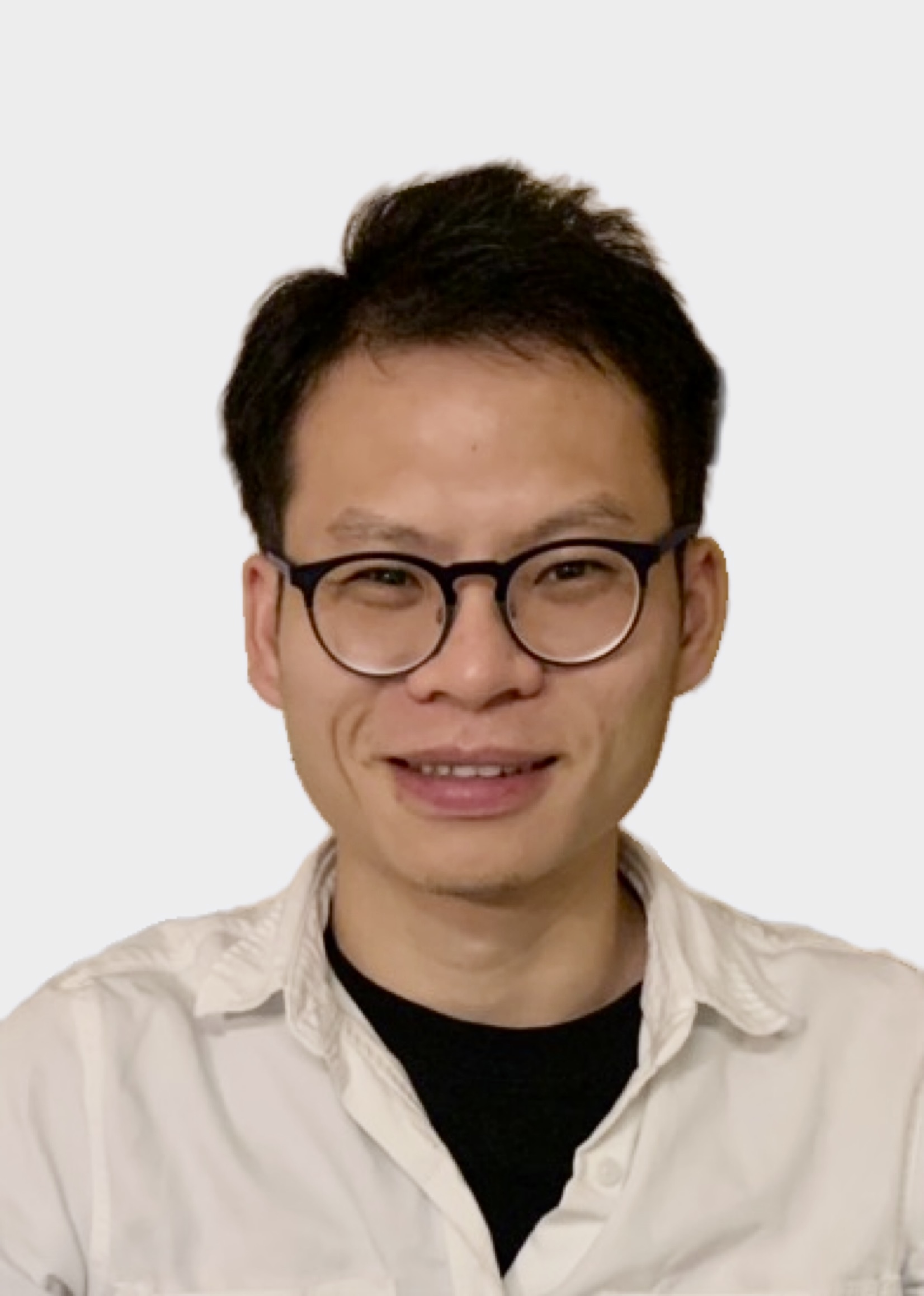}}]{Shenghui Li} received the B.S. degree from Xidian University, Xi'an, China, in 2017, and the M.S. degree from Sun Yat-sen University, Guangzhou, China, in 2019. He is currently pursuing the Ph.D. degree with the Department of Information Technology, Uppsala University, Uppsala, Sweden.
		
	His current research interests include federated learning, distributed optimization, and statistical machine learning.
		
	\end{IEEEbiography}
	
	\begin{IEEEbiography}[{\includegraphics[width=1in,height=1.25in,clip,keepaspectratio]{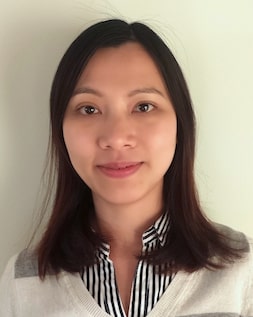}}]{Edith C.-H. Ngai}
		is currently an Associate Professor in Department of Electrical and Electronic Engineering, The University of Hong Kong. Before joining HKU in 2020, she was an Associate Professor in the Department of Information Technology, Uppsala University, Sweden. Dr. Ngai was a guest researcher at Ericsson Research Sweden in 2015-2017. Previously, she has conducted research in Imperial College London, Simon Fraser University, Tsinghua University, and UCLA. Her research interests include Internet-of-Things, machine learning, data analytics, and smart cities. Dr. Ngai is a VINNMER Fellow (2009) awarded by Swedish Governmental Research Funding Agency VINNOVA. She led the “Green IoT” project in Sweden, which was named on the IVA's 100-list from the Royal Swedish Academy of Engineering Sciences in 2020. She is currently an Area Editor of IEEE Internet of Things Journal and an Associate Editor of IEEE Access and IEEE Transactions of Industrial Informatics.
	\end{IEEEbiography}
	
	\begin{IEEEbiography}[{\includegraphics[width=1in,height=1.25in,clip,keepaspectratio]{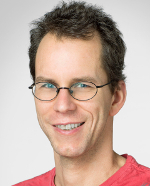}}]{Thiemo~Voigt}
		received the Ph.D. degree from Uppsala University, Sweden,
		in 2002. He is currently a Professor at the Department of Electrical Engineering at Uppsala University where he leads the Division of Networked Embedded Systems. He is also a senior researcher at RISE Computer Science. His current research interests include low-power networking, system software for embedded networked devices and the Internet of Things. His work has been cited more than 19000 times. He is a member of the editorial board for the IEEE IoT Newsletter and ACM Transactions on Sensor Networks (TOSN).
	\end{IEEEbiography}

\end{document}

%% file: includes.tex
\usepackage{nicefrac}       %
\usepackage{microtype}      %
\usepackage{caption}
\usepackage{subcaption}
\usepackage{comment}
\usepackage[switch]{lineno}
\usepackage{amsthm,amsfonts,amssymb}       %
\usepackage{thmtools, thm-restate}
\usepackage{times}
\usepackage{graphicx} %
\usepackage{color}
\usepackage{wrapfig,lipsum,booktabs}
\usepackage{array}
\usepackage{adjustbox}
\usepackage{multirow}
\usepackage{soul}
\usepackage{amsmath,bm}
\usepackage{xspace}
\usepackage{transparent}
\usepackage{fancyhdr}
\usepackage{makecell}
\usepackage{url}
\usepackage[hidelinks]{hyperref}

\usepackage[utf8]{inputenc}
\usepackage[T1]{fontenc}
\usepackage[outline]{contour}
\usepackage{color,soul}

\usepackage{algorithm}
\usepackage{algpseudocode}
\renewcommand{\algorithmiccomment}[1]{\bgroup\hfill\small//~#1\egroup}

\newcolumntype{H}{>{\setbox0=\hbox\bgroup}c<{\egroup}@{}}
\newcommand{\noaistats}[1]{}  %

\definecolor{darkgreen}{rgb}{0,0.4,0.0}
\definecolor{darkblue}{rgb}{0,0.1,0.3}
\definecolor{darkred}{rgb}{0.7,0.0,0.0}

\newcommand{\loss}{\ell}

\newcommand{\algfont}[1]{\textit{#1}}
\newcommand{\aggfont}[1]{\texttt{#1}}

\DeclareMathAlphabet{\mathpzc}{OT1}{pzc}{m}{it}

\newcommand{\bw}{\bm{w}}
\newcommand{\bz}{{\boldsymbol{z}}}

\newcommand{\update}{\bm{\Delta}}
\newcommand{\unit}{\bm{e}}

\newcommand{\grad}{\triangledown}

\DeclareMathOperator*{\argmin}{argmin}

\newcommand{\fedavg}{\algfont{FedAvg}\xspace}
\newcommand{\fedsgd}{\algfont{FedSGD}\xspace}
\newcommand{\syncsgd}{\algfont{sync-SGD}\xspace}

\newcommand{\ours}{\aggfont{ClippedClustering}\xspace}
\newcommand{\clustering}{\aggfont{Clustering}\xspace}
\newcommand{\gm}{\aggfont{GeoMed}\xspace}
\newcommand{\mean}{\aggfont{Mean}\xspace}
\newcommand{\median}{\aggfont{Median}\xspace}
\newcommand{\cc}{\aggfont{CC}\xspace}
\newcommand{\krum}{\aggfont{Krum}\xspace}
\newcommand{\tm}{\aggfont{TrimmedMean}\xspace}
\newcommand{\autogm}{\aggfont{AutoGM}\xspace}

\usepackage{pgf}
\usepackage{pifont}
\usepgflibrary{arrows}
\usepgflibrary{shapes}

\usepackage{xcolor} 
\usepackage{tikz}
\usetikzlibrary{fit,calc}

\colorlet{yellow}{yellow}

%% file: defense.tex
 \section{Aggregation Schemes for Evaluation}
 \label{aggs}
 In this section, we survey existing robust aggregation schemes, which represent state-of-the-art methods in the literature. Then, we propose a new scheme \ours, which addresses the weakness of the clustering-based scheme. Other than that, we provide a taxonomy of the eight aggregation schemes evaluated in our experiments.

All aggregation schemes considered in this paper are working on each round separately. For the sake of readability, we will omit the notation of the round $^t$ in the following sections. 
 	
 \subsection{Krum}
 \krum~\cite{blanchard2017machine} strives to find one of the local model updates that is closest to another $K-M-2$ ones with respect to squared Euclidean distance, which can be expressed by:
 \begin{align}
	\small
	\notag Krum := \{\update_i | i = \argmin\limits_{i \in [K]} \sum_{i \rightarrow j}  \lVert \update_i - \update_j \rVert^2 \}, 
\end{align} where $i \rightarrow j$ is the indices of the $K-M-2$ nearest neighbours of $\update_i$ measured by squared Euclidean distance, recall that $K$ is the number of clients in total, and $M$ is the number of malicious clients.

Under the \fedsgd framework, \krum was proven to converge with an important assumption that $c_1 \sigma < \lVert g \rVert$, where $c_1$ is a constant factor depending on the number of malicious clients and the dimension of model parameters, $\sigma$ is the maximal variance of the updates and $\lVert g \rVert$ is the expectation of updates.

\subsection{GeoMed}

The Geometric Median (\gm)~\cite{chen2017distributed,pillutla2022robust } scheme aims to find a vector that minimizes the sum of its Euclidean distances to all the update vectors:
    \begin{align}
        \label{eq_gm}
        GeoMed := \argmin_{\bz}  \sum_{k \in [K]} \lVert \bz - \update_k \rVert.
    \end{align} Although there is no closed-form solution to the \gm problem, a $(1 + \epsilon)$-approximate solution can be computed in nearly linear time~\cite{cohen2016geometric}.

Similar to \krum, \gm was also proven to converge under the \fedsgd framework, with the assumption that $c_2 \sigma < \lVert g \rVert$, where $c_2$ is a another constant factor that differs from $c_1$.

\subsection{AutoGM}

Auto-weighted Geometric Median (\autogm)~\cite{li2021byzantine} is a generalized version of \gm. \autogm aggregates the updates by solving the following problem:
    \begin{align}
        \label{eq_weighted_gm}
        AutoGM :=&   \argmin_{\bz}  \sum_{k \in [K]}\alpha_k \lVert \bz - {\update}_k \rVert + \frac{\lambda}{2}  \lVert \bm{\alpha}\rVert^2,  \\
        \notag s.t. \quad & \bm{\alpha} \in \mathbb{R}_+^K, \bm{1}^{\top}\bm{\alpha} = 1,
    \end{align}
    where 
    $\lambda$ is a user-specified hyper-parameter that controls the smoothness of $\bm{\alpha}$.

The key idea of optimizing \autogm is to divide the problem into two parts, i.e., one subproblem for estimating the weighted \gm, and the other subproblem for weighting the importance of each point. Then, we can minimize the objective iteratively with respect to one variable each time while fixing the other one~\cite{li2021byzantine}.

\subsection{Median}
\median~\cite{yin2018byzantine} is defined as the coordinate-wise median of the given set of updates, i.e., 
\begin{align}
	\notag med := Median&( \{ \update_k : k \in [K] \} ),
\end{align}
where the $i$-th coordinate $med_i = median(  \{ \update_k^i : k \in [K] \} )$, and $median$ is the usual (one-dimensional) median.

When using the \fedsgd framework, the robustness of the \median scheme is based on the assumptions that the gradient of the loss function has bounded variance, and each coordinate of the gradient has coordinate-wise bounded absolute skewness~\cite{yin2018byzantine}.

\subsection{TrimmedMean} 
The \tm~\cite{yin2018byzantine} aggregation scheme computes the coordinate-wise trimmed average of the model updates, which can be expressed by:
\begin{align}
	\notag trmean := TrimmedMean( \{ \update_k : k \in [K] \} ),
\end{align}
where the $i$-th coordinate $trmean_i = \frac{1}{(1-2\beta)m} \sum_{x \in U_k}x$, and $U_k$ is a subset obtained by removing the largest and smallest $\beta$ fraction of its elements.

In addition to the aforementioned assumptions for \median, the robustness of \tm relies on one stronger assumption that all the moments
of the derivatives of the loss function are bounded~\cite{yin2018byzantine}.

\subsection{Centered Clipping (\cc)} 
Centered Clipping (\cc)~\cite{karimireddy2021learning} iteratively clips the updates around the center while updating the center accordingly. For $l \ge 0$, \cc computes
\begin{equation} 
  \Delta_{l+1} \leftarrow \Delta_{l} + \frac{1}{K} \sum_{k \in [K]} (\Delta_i - \Delta_{l}) \min(1, \frac{\tau_1}{\lVert \Delta_i - \Delta_l \rVert}),
\end{equation} where $\Delta_{0}$ is assigned with the aggregated updates in the  previous round.

Karimireddy et al.~\cite{karimireddy2021learning} proofed the robustness of the \cc scheme when the variance of updates is bounded and $\frac{K}{M} \leq 0.15$.

\subsection{Clustering} 
This \clustering aggregation scheme~\cite{sattler2020byzantine,lisp2021byzantine} first calculates the pairwise cosine distances between their parameter updates, i.e., 
\begin{equation}
	\alpha_{i,j} := 1 - \frac{ \langle \update_i, \update_j \rangle}{\lVert \update_i \rVert \lVert \update_j \rVert}, 
\end{equation} then it separates the client population into two groups based on the cosine similarities using agglomerative clustering with average linkage. Finally, it aggregates the updates in the largest group using \mean.

Despite the lack of theoretical guarantee of robustness, \clustering achieves superior robustness in some cases, as we will show in Section~\ref{evaluation}. However, an obvious drawback of clustering using cosine similarities is that it only considers the relative directions, ignoring the magnitude of each vector. The attackers thus can fool the clustering scheme by simply amplifying the updates without changing their directions. As a consequence, the resulting updates added to the parameters will make the model jump over the minima and prevent the convergence of the optimization without being detected.

\subsection{ClippedClustering} 
We enhance the robustness of the aforementioned \clustering aggregation scheme by performing a clipping on all the updates before clustering, i.e., 
\begin{equation}
  \Delta_{k} \leftarrow \Delta_{k}\min(1, \frac{\tau}{\lVert \Delta_k \rVert}).
\end{equation} Here, $\tau$ is a clipping value hyper-parameter that is determined by the server. 
Note that this clipping scheme is the so-called clip by norm, not clip by value, where individual values of the update vectors are clipped if they go beyond a pre-set value. In clip by norm, the entire update is scaled if the norm of the update exceeds the threshold $\tau$. Thus we place a maximum on the magnitude of each vector that can be taken during training, preventing the attackers from amplifying the updates in the same direction. If the norm of update is below the threshold $\tau$, the update is unaffected.

Inspired by~\cite{seetharaman2020autoclip}, we design an automatic clipping strategy to defend against potential amplified malicious updates that the naive cosine similarity-based clustering scheme cannot handle well. Specifically, we set the clipping value hyper-parameter based on the statistics of the historical norms of the updates uploaded during training, i.e., we save the update norms up to current iteration and automatically set $\tau$ using the $50$-th percentile value (i.e., the median) of the history.

There are two reasons of choosing the median update norm as the clipping threshold: First, prior work~\cite{andrew2021differentially} has demonstrated that adaptive clipping to median norm works well across a range of general federated learning tasks without the need to tune any clipping hyper-parameter. Second, the median itself is a robust statistical measure of central tendency. As we assume that the majority of the clients are benign, malicious clients are unable to control the median norm even if they have full knowledge of all the updates. 

	\newcommand \rot{81}
	\newcommand{\rotbox}[1]{\rotatebox[origin=c]{\rot}{#1}}
	\begin{table}[t!]
		\centering
		{	
			\renewcommand{\arraystretch}{1.2} 
			\begin{tabular}{ |c||c|c|c|c|c|}
				\hline
				Defense                                & \rotbox{Euclidean distance} & \rotbox{Mean} & \rotbox{Median} & \rotbox{Cosine similarity} & \rotbox{Clipping} \\ \hline
				\krum~\cite{blanchard2017machine}       &         \checkmark          &                     &                       &                            &                   \\ \hline
				\gm~\cite{chen2017distributed}      &         \checkmark          &                     &      \checkmark       &                            &                   \\ \hline
				\autogm~\cite{li2021byzantine}          &         \checkmark          &                     &      \checkmark       &                            &         \\ \hline
				\median~\cite{yin2018byzantine}         &                             &                     &      \checkmark       &                            &                   \\ \hline
				\tm~\cite{yin2018byzantine}   &                             &     \checkmark      &                       &                            &                    \\ \hline
				\cc~\cite{karimireddy2021learning}      &                             &     \checkmark      &                       &                            &    \checkmark     \\ \hline
				\clustering~\cite{sattler2020byzantine,lisp2021byzantine} &                             &     \checkmark      &                       &         \checkmark         &                   \\ \hline
				\ours~(ours)                      &                             &     \checkmark      &                       &         \checkmark         &    \checkmark     \\ \hline
			\end{tabular}
		}
		\caption{Summary of the robust aggregation schemes and their main defense mechanisms.}
		\label{taxonimy}
	\end{table}

\subsection{Taxonomy}

Table~\ref{taxonimy} shows a taxonomy of the eight aggregation schemes, where \krum, \gm, and \autogm are typical Euclidean distance-based schemes, i.e., they are all designed to find a vector closest to the updates measured by Euclidean distance. Among them, \gm and \autogm are both based on geometric median. The \median scheme simply computes the coordinate-wise median instead of geometric median. \tm, \cc, \clustering and \ours are all categorized as mean-based schemes as they eventually compute the mean, although they also utilize other mechanisms. \clustering and \ours both perform clustering based on cosine similarity while \ours clips the updates before clustering.

\begin{figure}
	\centering
	\includegraphics[width=0.6\linewidth]{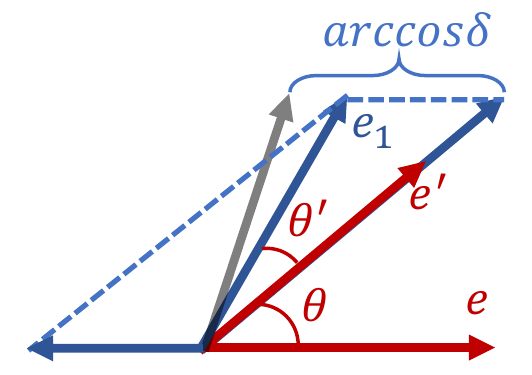}
	\caption{Demonstration of the solution to \eqref{eq_unit_problem}. The solution of $\unit_1$ can be expressed as a linear combination of $\unit$ and $\unit^\prime$.}
	\label{fig_adaptive_sol}
\end{figure}

\section{Adaptive Attack on \ours}
\label{sec_adp}
In this section, we design a strong adaptive attack on \ours. We assume that the attacker has full knowledge of the system, including the aggregation scheme and all the updates from benign clients.

Given the fact that the updates are clipped by the server if their magnitudes are beyond the threshold, state-of-the-art approach~\cite{shejwalkar2021manipulating} that aims to maximize Euclidean distance between the aggregated vector and the true update is no longer applicable. Instead, the attack would be more effective if it can change the direction of update without being excluded by the aggregation scheme.

The idea of our attack is to make sure that all the malicious updates stay in the largest cluster, while deviating from the correct direction as possible. Since the clustering method and linkage function are known to us, we can perform the same clustering process with the benign updates only. Then we can carefully design malicious updates to make it close enough to the largest cluster, without breaking the existing structure. 
Specifically, we compute the average cosine similarity of the two benign clusters, denoted by $\delta$, and use it as the bound to design malicious updates. 

For simplification, we let $\update_1 = \dots = \update_M$, and denote by $\unit_i = \frac{\update_i}{\lVert \update_i \rVert} $ the unit vector whose direction is the same as $\update_i$. The problem becomes:
\begin{align}
        \label{eq_unit_problem}
        \min_{\unit_1}  \quad & \langle \unit_1 , \unit \rangle,  \\
        \notag s.t. \quad & \langle \unit_1 , \unit^\prime \rangle > \delta,
\end{align}
where $\unit$ is the unit vector of $\update$, and $\unit^\prime$ is the unit vector of the center of the largest benign cluster. The constraint ensures that the malicious group is included by the largest cluster when we perform hierarchical clustering based on average linkage.

Minimizing the problem is equivalent to maximizing the angle between $\unit_1$ and $\unit$, with the constraint that the angle between $\unit_1$ and $\unit^\prime$ should be smaller than $\arccos{\delta}$, which is demonstrated in Fig.~\ref{fig_adaptive_sol}. Now we are able to solve the problem using $\{\unit, \unit^\prime\}$ a basis, i.e., let \begin{align}
    \theta &= \arccos{\langle \unit , \unit^\prime \rangle},  \\
    \theta^\prime &= \arccos \delta - \epsilon,
\end{align} where $\epsilon > 0$ is a small enough number. The solution to \eqref{eq_unit_problem} is expressed as: 
\begin{equation}
    \unit_1 = \left( \cos (\theta + \theta^\prime) - \frac{\sin (\theta + \theta^\prime)}{\tan \theta} \right) \unit + \left( \cos \theta^\prime + \frac{\sin \theta^\prime}{\tan \theta} \right) \unit^\prime.
\end{equation}

Once we obtain the unit vector of malicious updates, we can scale the magnitude by the clipping threshold $\tau$, i.e., 
\begin{equation}
    \update_1 = \dots = \update_M = \tau \unit_1.
\end{equation}

We note that such an attack is applicable to agglomerative clustering with average linkage. For complete linkage, one can simply replace $\delta$ with the minimum cosine similarity of the two benign clusters and solve the problem in the same manner. Other linkage functions are omitted due to space limitations. 

\begin{figure}[t]
	\centering
	\includegraphics[width=\linewidth]{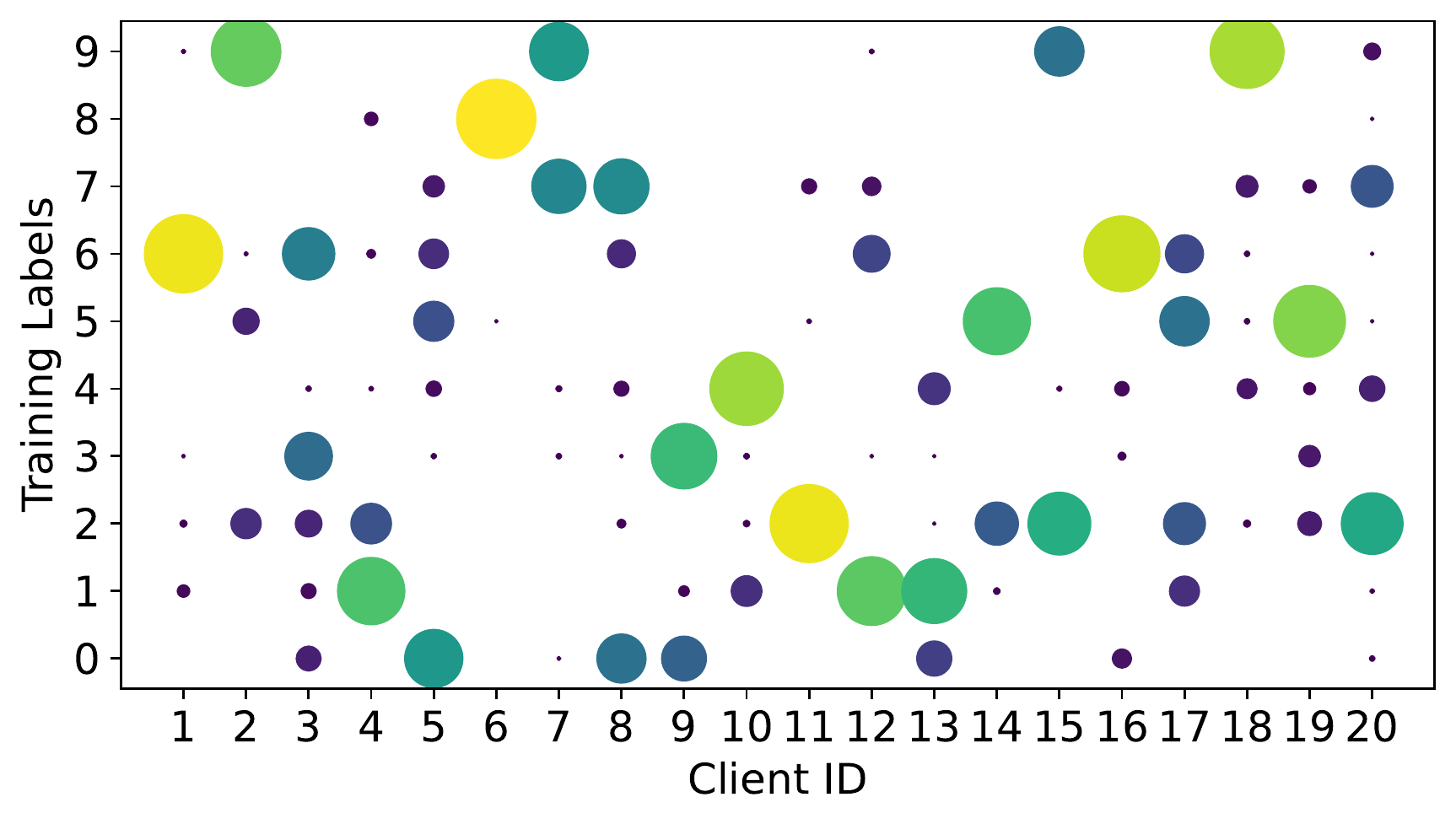}
	\caption{Visualization of statistical heterogeneity among clients on our Non-IID partition of CIFAR-10 dataset, where the $x$-axis indicates client IDs, the $y$-axis indicates class labels, and the size of scattered points indicates the number of training samples for a label available to that client.}
	\label{fig_noniid}
\end{figure}

	\begin{figure*}[ht]
	    \centering
	    \includegraphics[width=\linewidth]{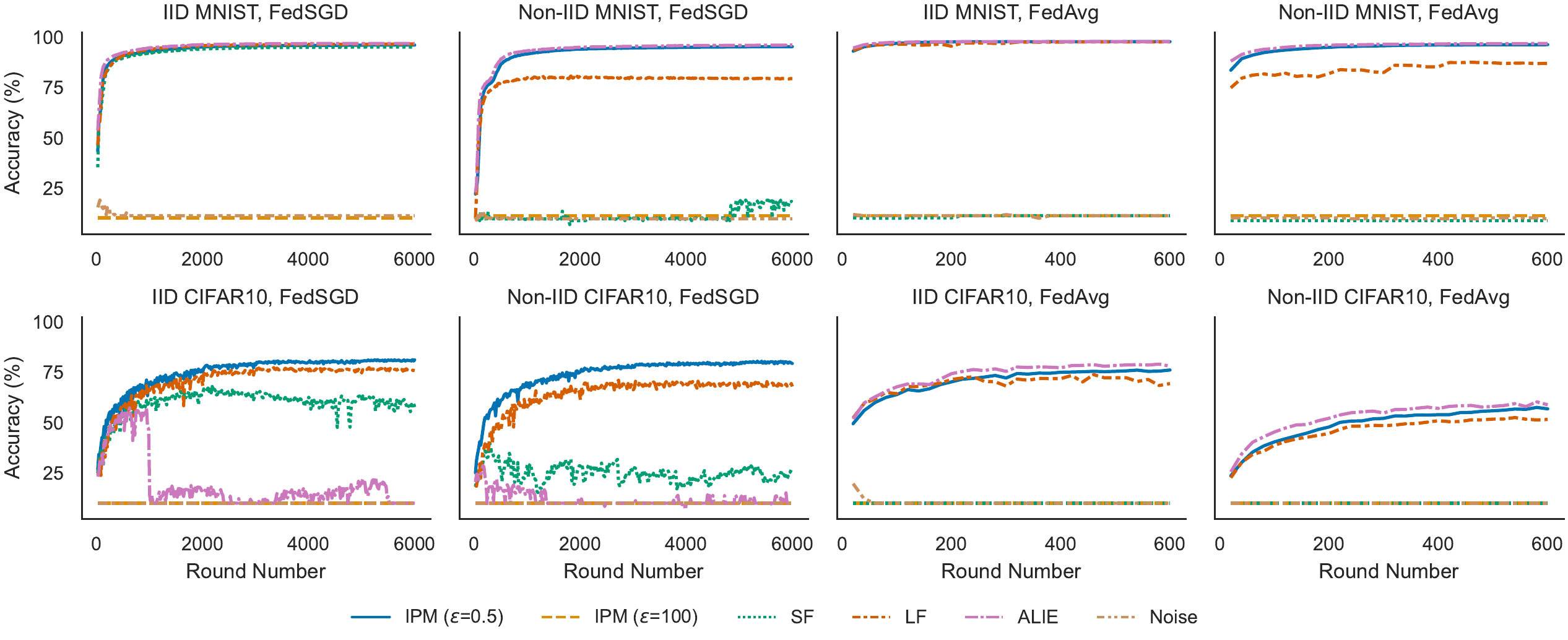}
	    \caption{Impact of malicious attacks on the \mean aggregation scheme. Overall, compared with using IID datasets, training with Non-IID datasets is more vulnerable to attacks.}
	    \label{fig_mean_agg}
	\end{figure*}

%% file: related_work.tex

\section{Related Work}
\label{relatedwork}
\subsection{Byzantine Attacks on FL}
Byzantine attacks on FL are carried out by malicious clients during the distributed optimization of machine learning models, aiming to bias the global model to the desire of malicious clients~\cite{lyu2020threats}. Depending on the adversarial goals, Byzantine attacks in FL can be classified into two categories: targeted attacks and untargeted attacks~\cite{jere2020taxonomy,lyu2020threats}. Targeted attacks, such as backdoor attacks, aim to make the global model generate attacker-desired misclassifications for some particular test samples~\cite{xie2019dba,bagdasaryan2020backdoor,andreina2021baffle}. While untargeted attacks aim to degrade the overall performance of the global model indiscriminately~\cite{fang2020local}.

We particularly focus on untargeted attacks as most Byzantine-robust studies do~\cite{blanchard2017machine,yin2018byzantine,chen2017distributed,li2021byzantine,karimireddy2021learning,sattler2020byzantine}. Many studies considered to launch attacks by adding Gaussian noise or flipping the sign of the actual updates~\cite{li2021byzantine,li2019rsa}. Those attacks, however, can be detected by some Euclidean distance-based aggregation schemes such as \krum~\cite{blanchard2017machine}, as they usually make the malicious updates far from the benign ones when measured by Euclidean distance. On the other hand, Baruch et al. \cite{baruch2019little} showed that the attackers can actually circumvent robust schemes including \tm and \krum by taking advantage of empirical variance between the updates of clients if such variance is high enough. Furthermore, Xie et al. proposed an attack strategy, Inner Product Manipulation (IPM) attack that poses a significant threat to \median and \krum by manipulating the inner product between the true mean of the updates and the output of the aggregation schemes~\cite{xie2020fall}.

There exists a debate over Byzantine attacks in real-world FL systems. Shejwalkar et al.~\cite{shejwalkar2022back} claimed that the fraction of compromised genuine clients are usually small (e.g., 0.01\%) in practice due to the high cost of hacking and manipulating multiple devices simultaneously, resulting in a very limited impact on the global model. However, Cao et al.~\cite{cao2022mpaf} argued that the attackers may address this limitation by injecting fake clients into FL systems using zombie devices and simulators. 

\subsection{Byzantine-robust FL}
In FL settings, a number of strategies have been explored to defend against specific types of attacks or failures, including backdoor attacks~\cite{andreina2021baffle,zhao2021federatedreverse}, free-rider attacks~\cite{lin2019free,fraboni2021free}, and gradient inversion attacks~\cite{huang2021evaluating,geiping2020inverting}. On a more general level, Byzantine-robust FL solutions aim to mitigate the effect of arbitrary updates uploaded by malicious clients, instead of focusing on specific types of attacks~\cite{fang2020local}. Those Byzantine-robust solutions can be classified into three categories: redundancy-based schemes, trust-based schemes, and robust aggregation schemes.

Redundancy-based schemes assign each client redundant updates and use this redundancy to eliminate the effect of Byzantine failures~\cite{chen2018draco,data2018data,rajput2019detox}. 
In 2018, Chen et al.~\cite{chen2018draco} presented a framework, DRACO, for robust distributed training that uses ideas from coding theory. In DRACO, each client evaluates redundant gradients that are used by the server to eliminate the effects of adversarial updates. In 2019, Rajput et al. presented DETOX, a framework that combines algorithmic redundancy with robust aggregation. The defense of DETOX operates in two steps, a filtering step that uses limited redundancy to significantly reduce the effect of Byzantine nodes, and a hierarchical aggregation step that can be used in tandem with any state-of-the-art robust aggregation method. However, these redundant updates, in the worst case, require each node to compute $\Omega(M)$ times more updates, where $M$ is the number of Byzantine clients~\cite{chen2018draco}. This overhead is prohibitive in settings with a large number of Byzantine clients. In 2021, Cao et al.~\cite{cao2021provably} proposed to use a randomly selected subset of clients to learn redundant global models. At inference time, it takes the majority vote among the global models when predicting the label of a testing sample. The authors showed that such an ensemble approach with any base FL algorithm is provably secure against malicious clients.

Trust-based schemes assume that some of the clients or datasets are trusted for filtering and re-weighting the local model updates\cite{li2019rsa,konstantinov2019robust, park2021sageflow,cao2021fltrust}. For example, in 2019, Li et al.~\cite{li2019rsa} proposed to incorporate the objective function with a regularization term, which minimizes the distance between the server parameters and the client parameters. In 2021, Park et al.~\cite{park2021sageflow} designed an entropy-based filtering scheme to detect the outlier updates based on some trusted public data on the server side. During the training, the server computes the entropy of each update with the trusted dataset. Based on their experimental observations, they argue that the updates with higher entropy will lead to lower accuracy during the testing stage. Thus, they set a threshold for the entropy and filter out updates with entropy higher than the threshold. In 2021, Cao et al.~\cite{cao2021fltrust} utilized cosine similarity to measure the similarity between updates submitted by the clients and the update obtained by training based on the trusted dataset owned by the server. The authors argued that an attacker can manipulate the directions of updates to perform model poisoning attacks, and the directions of the updates can, to a certain extent, indicate the honesty of the end devices. After the calculation of cosine similarity, the server calculates a trust score for each update using the ReLu function. The score is then used as the weight for the global model aggregation. In general, trust-based schemes have the potential to deal with situations where more than half of the updates are malicious according to some pre-validated information to detect malicious updates. However, trusted datasets or clients are not always available for the server, for example, due to the concern of user data privacy.

Robust aggregation schemes estimate the global update based on the local updates according to their robust aggregation rules or algorithms~\cite{blanchard2017machine,yin2018byzantine,chen2017distributed,li2021byzantine,karimireddy2021learning,sattler2020byzantine}. Byzantine-robust aggregation has been explored to handle the devices sending corrupted updates to the server, including geometric median (\gm)~\cite{chen2017distributed}, \krum~\cite{blanchard2017machine}, \tm~\cite{yin2018byzantine}, and \median~\cite{yin2018byzantine}. They are commonly used to estimate the model parameters and mitigate the effect of malicious updates in global aggregation. In 2017, Chen et al.~\cite{chen2017distributed} proposed a \gm-based method to aggregate the gradients for distributed statistical machine learning and showed the robustness and convergence in i.i.d settings. In 2020, Wu et al.~\cite{wu2020federated} showed that \gm scheme provably provides improved Byzantine robustness compared to other aggregation schemes in FL. In 2022, Pillutla et al.~\cite{pillutla2022robust} applied \gm as a robust aggregation rule for FL and analyze the convergence of the resulting FL algorithm for least-squares objective with IID local datasets. In 2022, Li et al.~\cite{li2021byzantine} proposed \autogm, a variant of \gm that automatically re-scales the weight of each parameter component according to a user-specified threshold of skewness. According to our empirical study in this paper, this category of schemes all show limitations in \fedsgd and \fedavg algorithms in terms of Byzantine robustness.